\documentclass[11pt,a4paper]{article}
\usepackage{times,latexsym}
\usepackage{url}
\usepackage[T1]{fontenc}

\usepackage[acceptedWithA]{tacl2021v1}

\usepackage{xspace,mfirstuc,tabulary}

\newif\iftaclinstructions
\taclinstructionsfalse 
\iftaclinstructions
\renewcommand{\confidential}{}
\renewcommand{\anonsubtext}{(No author info supplied here, for consistency with
TACL-submission anonymization requirements)}
\newcommand{\instr}
\fi

\iftaclpubformat 

\else

\fi

\usepackage{graphicx}
\usepackage{xcolor}

\usepackage{amsmath}
\usepackage{amssymb}

\usepackage{subcaption}

\usepackage{enumitem}

\usepackage{multirow, makecell}

\usepackage{listings}
\lstnewenvironment{code}[1][]%
  {\noindent\minipage{\linewidth}\medskip 
   \lstset{basicstyle=\ttfamily,frame=single,#1}}
  {\endminipage}

\DeclareMathOperator*{\softmax}{Softmax}

\newcommand{\pawel}[1]{}
\newcommand{\jerry}[1]{}
\newcommand{\jan}[1]{}
\newcommand{\yusuf}[1]{}

\newcommand{\ch}[1]{#1}

\title{Scaling Attention Head Analysis via Gradient-Based Attribution in Context-Aware Machine Translation}

\author{
  Paweł Mąka \and Yusuf Can Semerci \and Jan Scholtes \and Gerasimos Spanakis
  \\
  \ \\
  Department of Advanced Computing Sciences
  \\
  Maastricht University
  \\
  \texttt{\{pawel.maka, y.semerci, j.scholtes, jerry.spanakis\}}\\
  \texttt{@maastrichtuniversity.nl}
}

\date{}

\begin{document}
\maketitle
\begin{abstract}
In this paper, we introduce a gradient-based head attribution strategy where the Token-level Max-Margin loss is backpropagated to the attention maps. This framework enables a large-scale causal analysis of attention heads, making it suitable for LLMs. We evaluate our method on the task of disambiguation in Context-aware Machine Translation, where we analyze 50 phenomena across 4 models and 4 language directions. We empirically show the alignment of our method with the effects of increasing the attention scores of token-to-token relations on three models and two language directions, ensuring the robustness of our method. Our analysis reveals the presence of the "general-purpose" attention heads that improve the model's performance when attending to different relations. We find that the average attention a head assigns to a relation does not necessarily relate to the model's performance, which suggests that the models developed redundancies during training in terms of the head functions.
\end{abstract}

\section{Introduction}

\begin{figure*}
    \centering
    \includegraphics[width=1\linewidth]{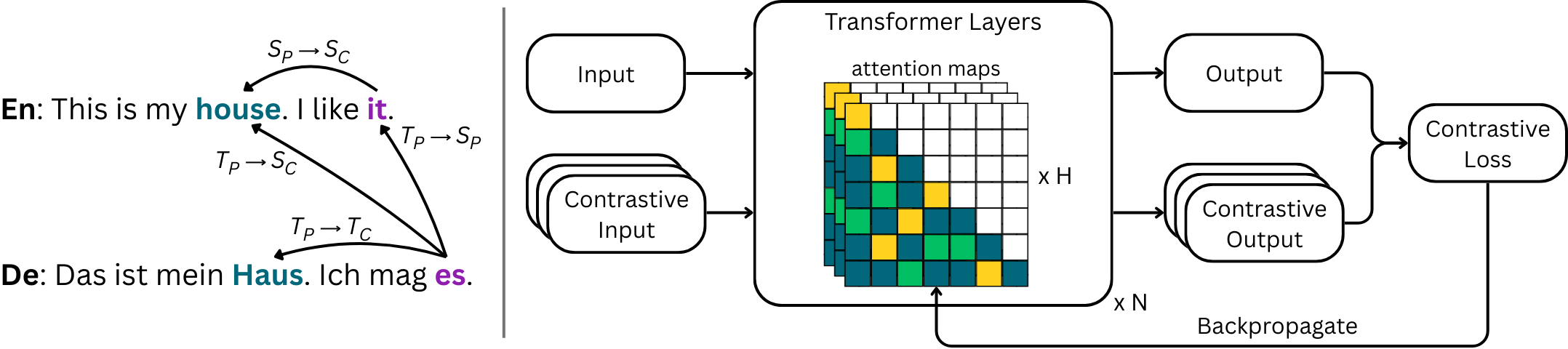}
    \caption{The token-to-token relations that we consider in this work (left). The illustration of the proposed strategy (right).}
    \label{fig:method}
\end{figure*}

In Context-Aware Machine Translation (MT), providing the system with surrounding sentences allows it to maintain document-level coherence and resolve linguistic ambiguities \citep{agrawal2018contextual, bawden-etal-2018-evaluating, voita-etal-2019-good}. While previous works often focused on specialized architectures \citep{tu-etal-2017-context, bawden-etal-2018-evaluating, miculicich-etal-2018-document, maruf-etal-2019-selective, huo-etal-2020-diving, zheng2021towards} or the encoder-decoder Transformer models, often referred to as the single-encoder architecture \citep{vaswani2017attention, sun-etal-2022-rethinking, majumde2022baseline, gete-etal-2023-works, post2023escaping, mohammed-niculae-2024-measuring}, the research focus has recently shifted toward the use of Large Language Models (LLMs) \citep{wang-etal-2023-document-level, wang-etal-2024-benchmarking, alves2024tower, cui-etal-2024-efficiently, wu2024adaptinglargelanguagemodels}, which utilize the decoder-only Transformer architecture to process context autoregressively. Despite the performance of these models \citep{pang-etal-2025-salute}, their size poses a challenge to analyzing their internal behavior.

Research on interpretability has demonstrated that specific attention heads within Transformers often learn to perform distinct, human-interpretable functions \citep{vig-belinkov-2019-analyzing, clark-etal-2019-bert, voita-etal-2019-analyzing, Olsson2022IncontextLA, li2023inference}. Techniques such as Activation Patching have been widely used to identify which components are causally responsible for specific model behaviors \citep{wang2023interpretability, zhang2025the}. Related to this, Modifying Heads \citep{maka-etal-2025-analyzing}, which is a technique where attention scores for specific linguistic relations are artificially shifted, has emerged as a tool to measure the causal importance of head-relation pairs. By forcing a head to attend to a specific contextual cue (e.g., a pronoun-antecedent relation), we can observe the impact on model performance as indicated by the accuracy in contrastive evaluations \citep{muller-etal-2018-large, lopes-etal-2020-document}.

Nevertheless, the computational cost of these causal interventions is significant because in modern LLMs with dozens of layers, there exist thousands of potential head-relation pairs. Consequently, performing a full sweep of modifications to identify functionally important components is prohibitively expensive. For instance, evaluating the EuroLLM 1.7B model \citep{martins2025eurollm} across all relevant relations requires thousands of individual evaluation runs. This limits our ability to scale the intervention-based studies to larger models or to include more language directions and linguistic phenomena.

In this paper, we propose an efficient framework that automates the head-relation attribution, which extends the scalability of the interpretability studies for LLMs. Our method approximates the effects of head modification using gradient attribution. By backpropagating a loss into the attention maps \ch{(not to model's parameters or inputs to the network)}, we can identify promising heads without the need to exhaustively evaluate each head-relation pair (see Figure~\ref{fig:method} for an overview). 
\ch{We introduce and evaluate the Token-level Max-Margin loss $\mathcal{L}_{TMM}$, designed to align the gradient signal with the change in the accuracy of the contrastive evaluation after attention intervention.}
This allows us to select only the most promising (top-$k$) head-relation pairs for modification, significantly reducing the computational cost while maintaining the robustness of our analysis. 
We conduct a large-scale evaluation by analyzing 50 contextual phenomena across four language directions in four open-weights instruction-tuned LLMs: EuroLLM 1.7B \citep{martins2025eurollm} and 9B \citep{ramos2026eurollm22btechnicalreport}, Qwen 2.5 1.5B \citep{qwen2, qwen2_5}, and Gemma 3 1B \citep{gemma_2025}.
We study the following contextual phenomena: Gender, Formality, Auxiliary, and Inflection (see Appendix~\ref{sec:contrastive-ctxpro} for more details).
\ch{While we concentrate our evaluation on the task of Context-aware MT, the technique can, in principle, be applied to any task where a plausible token-to-token relation can be defined, which we test on the task of Indirect Object Identification~\citep{wang2023interpretability}.}

Our contributions are as follows:
\begin{itemize}[topsep=0pt,itemsep=0pt,partopsep=0pt, parsep=0pt, itemindent=15pt, leftmargin=0pt]
    \item We propose a gradient-attribution framework to automatically identify salient head-relation pairs responsible for context utilization.
    \item We demonstrate that focusing on top-$k$ heads identified via gradients preserves the accuracy of the analysis while reducing computational requirements by more than 96\%.
    \item We perform a large-scale analysis of four LLMs across four language directions. We confirm the presence of attention heads in LLMs that not only attend to linguistically relevant token-to-token relations but also improve the performance when the attention is increased.
    \item We discover attention heads that exhibit multi-functional behavior (attending to multiple specific relations), and we link this overlap with the generalizability of the models, which could open possibilities for improved transfer.
    \item We demonstrate a dissociation between the average attention score and the model's performance, suggesting that LLMs develop functional redundancies. 
\end{itemize}

\section{Related Work}

Historically, many dedicated architectures have been proposed for Context-aware MT \citep{miculicich-etal-2018-document, voita-etal-2019-good, voita-etal-2019-context, bao-etal-2021-g, chen2022one, feng-etal-2022-learn, bulatov2022recurrent} including the popular multi-encoder in which a separate encoder is responsible for processing the context sentences \citep{jean2017does, miculicich-etal-2018-document, maruf-etal-2019-selective, huo-etal-2020-diving, zheng2021towards}, but the standard Transformer model with the sentences being concatenated, known as single-encoder \citep{vaswani2017attention, tiedemann-scherrer-2017-neural, ma-etal-2020-simple, zhang-etal-2020-long}, exhibited high performance despite its relative simplicity \citep{majumde2022baseline, sun-etal-2022-rethinking, gete-etal-2023-works, post2023escaping}. Because of model scaling and pretraining on massive-scale datasets, the decoder-only LLMs have achieved state-of-the-art results in Context-aware MT \citep{wang-etal-2023-document-level, wang-etal-2024-benchmarking, alves2024tower, cui-etal-2024-efficiently, kocmi-etal-2024-findings, wu2024adaptinglargelanguagemodels, pang-etal-2025-salute}. However, the scale of the models presents a challenge for analyzing how the internal mechanisms in the models relate to their performance on the end task.

Research has shown that the general-purpose translation metrics, such as BLEU \citep{papineni-etal-2002-bleu} and COMET \citep{rei-etal-2020-comet}, do not measure how well the models use context \citep{hardmeier2012discourse, wong-kit-2012-extending}. This is because often only a few tokens require context in a given sentence (e.g., pronouns in a gendered language), and the standard metrics are insensitive to such small mistakes. To mitigate this issue, both contrastive \citep{muller-etal-2018-large, bawden-etal-2018-evaluating, voita-etal-2019-good, lopes-etal-2020-document} and generative \citep{wicks-post-2023-identifying} evaluation datasets have been introduced.

Numerous methods aimed at explaining Natural Language Processing models have been proposed \citep{bau2019identifying, toneva2019interpreting, ferrando-etal-2022-towards, langedijk-etal-2024-decoderlens, meng2024locating}. While using raw attention scores as explanations is a topic of debate \citep{jain-wallace-2019-attention, wiegreffe-pinter-2019-attention}, attention remains the primary mechanism in the Transformer model enabling contextualization. Therefore, understanding its influence on the model's behavior is an active area of research \citep{wiegreffe-pinter-2019-attention, abnar-zuidema-2020-quantifying, kobayashi-etal-2020-attention, kobayashi-etal-2021-incorporating, bogoychev-2021-parameters, gheini-etal-2021-cross, mohebbi-etal-2023-quantifying}. Several works suggest that some heads exhibit specific functions \citep{clark-etal-2019-bert, voita-etal-2019-analyzing, Olsson2022IncontextLA, todd2024function} often linked to linguistically plausible relations \citep{vig-belinkov-2019-analyzing, tenney-etal-2019-bert, jo-myaeng-2020-roles}.

Similar to our approach, many works on interpretability employed gradients to explain the decisions of the models \citep{bach2015pixel, pmlr-v70-shrikumar17a, 10.5555/3305890.3306024, yin-neubig-2022-interpreting} with attention heads being the focus of some of them \citep{8237336, Chefer_2021_CVPR, song-etal-2024-better, 11084559}. 
The ultimate goal of those works was to explain the models by identifying the parts of the inputs that influence the model's behavior (local explanations). Conversely, our work concentrates on identifying parts of the model itself (i.e., attention heads) that are influencing the final prediction, known as global explanation \citep{danilevsky-etal-2020-survey, madsen2022posthoc}.

Attention Patching \citep{wang2023interpretability, zhang2025the} and Modifying Heads \citep{maka-etal-2025-analyzing} are the intervention-based interpretability techniques that aim to identify meaningful functions of attention heads.  
\ch{While Attention Patching replaces the attention map for a head while processing the contrastive input with the saved attention map of the correct input,}
the Modifying Heads technique changes the attention to match the specified value $C$ for a specific token-to-token relation, maintaining the distribution of attention for other tokens. 
It necessitates the specification of relations of interest, but allows for measuring the potentially positive effect of increased attention and enables the targeted fine-tuning of attention heads \citep{maka-etal-2025-analyzing}. 

\section{Background}

In this section, we describe the existing concepts upon which our work builds. We start by formulating the attention mechanism in the Transformer model \citep{vaswani2017attention}. Next, we present the attention intervention from the Modifying Heads method \citep{maka-etal-2025-analyzing}. Lastly, we describe the contrastive evaluation in Context-aware MT \citep{muller-etal-2018-large, lopes-etal-2020-document}.

\subsection{Attention Mechanism}

In the Transformer architecture, at each timestep, the multi-head self-attention is responsible for gathering the information from all input tokens. It consists of a specified number of heads that perform self-attention, and their outputs are concatenated and linearly projected to form the final output of the attention module. First, each head maps the input representations into the $Q$, $K$, and $V$ projections by a learned linear transformation. Second, the scaled dot product between $Q$s and $K$s is calculated:
\begin{equation} \label{eq:transformer-qk-pre-softmax}
\begin{aligned}
H^{l,h} = \frac{Q^{l,h} {K^{l,h}}^T}{\sqrt{d_k}}, \\
\end{aligned}
\end{equation}
where $\sqrt{d_k}$ is the scaling factor, and $d_k$ is the dimension of the $Q$ and $K$ representations. Next, the attention scores $S^{l,h}$ are obtained by applying the softmax function to $H^{l,h}$:
\begin{equation} \label{eq:transformer-qk-softmax}
\begin{aligned}
S^{l,h} = \softmax{\left( H^{l,h} \right)}. \\
\end{aligned}
\end{equation}
Lastly, the output of head $h$ in $l$-th layer is calculated as the sum of $V$ representations weighted by the attention scores.

\subsection{Modifying Heads}

Modifying Heads \citep{maka-etal-2025-analyzing} adjusts the attention scores corresponding to a specified token-to-token relation $\mathcal{X} \rightarrow \mathcal{Y}$. It makes the total attention score of each token in $\mathcal{X}$ given to $\mathcal{Y}$ tokens equal a desired value $C$ while not changing the pre-softmax attention scores of all other tokens. 
It is formulated as follows:
\begin{equation} \label{eq:modified-heads}
\begin{aligned}
&\tilde{H}^{l,h}_{i,j} = \log \biggl( \frac{C}{|\mathcal{Y}| (1-C)} \sum_{k \in Y \setminus \mathcal{Y}} \exp \left( H^{l,h}_{i,k} \right) \biggr) \\ 
&\forall i \in \mathcal{X}, j \in \mathcal{Y},
\end{aligned}
\end{equation}
where $\tilde{H}$ represents the updated pre-softmax attention scores, $H$ are the original pre-softmax scores, $k \in Y \setminus \mathcal{Y}$ are the attended tokens not present in the subset of interest $\mathcal{Y}$. \citet{maka-etal-2025-analyzing} modified the heads to different values of $C$ for relations from ambiguous words to context cues and observed the changes in the accuracy of the models in contrastive evaluation. Intuitively, the method reveals how the model reacts to the changes in the function of one of its components. There are three notions of a head's function: (1) the head is attending a relation, which can be measured by the average attention score; (2) the head attending a relation is necessary for the end task, measured by the drop of performance when the head is modified to $C=0.01$; (3) the head attending a relation is beneficial to the task, indicated by the increase in the performance on the end task when modified to $C=0.99$. In our work, we focus on the effects of Modifying Heads to $C=0.99$, as it can lead to real-world improvements in the models through the targeted fine-tuning \citet{maka-etal-2025-analyzing}.

\subsection{Contrastive Evaluation}

Contrastive evaluation assesses whether the model effectively uses the contextual information and is based on the contrastive datasets. It presents the model with the source sentence, source and target contexts, and several versions of the target sentence with only a single word being different in each of them. The examples are chosen to make the correct version dependent on the context. The model assigns the probability to each version, which is calculated from the model's outputs with the whole sequence (including the target sentence) as input. If the right version is assigned the highest probability, the model is said to be correct. The final score is the accuracy $A$ of the model over the whole dataset.

In our study, we use two contrastive datasets: ContraPro \citep{muller-etal-2018-large} and Large Contrastive Pronoun Testset (LCPT; \citealp{lopes-etal-2020-document}), which target pronoun disambiguation in English-to-German and English-to-French, respectively. To expand the analysis to different language phenomena, we adapt the ctxPro dataset \citep{wicks-post-2023-identifying} to contrastive evaluation. The original version is used for generative evaluation, meaning that the model generates the translation, and the script checks whether it contains the expected phrase. We generate the contrastive examples by replacing the expected word in the gold translation with a word sampled from all expected words for a particular linguistic phenomenon. Because this simple method does not ensure that the replacement word is morphologically plausible, we set the number of contrastive examples to maintain the difficulty of the test. We provide more details about the process in Appendix~\ref{sec:contrastive-ctxpro}.


\section{Approach}

Our goal is to approximate the effects of Modifying Heads to $C=0.99$ for a specified token-to-token relation $\mathcal{X} \rightarrow \mathcal{Y}$ on the accuracy in the contrastive evaluation by using gradient attribution. For each example $d$ in the dataset $\mathcal{D}$, we perform the forward pass of both the correct $X^{c}$ and all incorrect $X^{w_i}$ inputs. Next, we calculate the loss $\mathcal{L}$ based on the outputs of the model and backpropagate it into the attention scores $S^{l, h}$ of each head in each layer\footnote{It should be noted that we do not use the gradients of the model's parameters but of the attention maps directly.}. We extract the gradients of the attention scores corresponding to the specified relation and aggregate them over the dataset:
\begin{equation} \label{eq:gradient-formulation}
\begin{aligned}
&G^{l,h}_{\mathcal{X} \rightarrow \mathcal{Y}} = \frac{1}{|\mathcal{D}|} \sum_{d \in \mathcal{D}} \max_{i \in \mathcal{X}^d, j \in \mathcal{Y}^d} \biggl(- \Bigl( \nabla_{S^{l,h}_c} \mathcal{L}^d \Bigr)_{i j} \biggr),
\end{aligned}
\end{equation}
where $\nabla_{S^{l,h}_c}$ represents the gradient with respect to the attention scores $S^{l,h}_c$ of the correct contrastive example, and $\mathcal{L}^d$ is the loss of the $d$-th example from the contrastive dataset $\mathcal{D}$. 
\ch{For the multi-token phrases and contextual cues, we aggregate the maximum gradient over the token pairs, as this functionally corresponds to a head attending any part of the cue from any part of the phrase.}

The choice of the loss function $\mathcal{L}$ determines the quality of the resulting gradients, as we explore in Section~\ref{sec:gradient-alignment}. Motivated by the contrastive evaluation, we propose to use max-margin loss based on contrastive tokens, which we call Token-level Max-Margin loss (TMM):
\begin{equation} \label{eq:token-max-margin-loss}
\begin{aligned}
& \mathcal{L}^d_{TMM} = \\& \max \biggl(0, \mu - \bar{P_\theta} \bigl( x^{d,c} \bigr) + \max_i \Bigl( \bar{P_\theta} \bigl( x^{d,w_i} \bigr) \Bigr) \biggr),
\end{aligned}
\end{equation}
where $\bar{P}_\theta (x)$ is the probability of the token $x$ assigned by the model, and margin $\mu$ is the hyperparameter. We select the lowest probability in case of multi-token contrastive words or phrases.
This loss guides the model's attention scores to increase the relative probability difference between contrastive tokens only in cases where the model is wrong (assigns lower probability to the correct token) or the difference is smaller than $\mu$. 

While the backpropagation of the loss into the attention scores of each head is a good indicator of the difference in accuracy $\Delta A$ on the contrastive dataset when modified to $C=0.99$, gradients are non-linear or even non-monotonic (see Section~\ref{sec:why-gradients-work}). For this reason, we still apply Modifying Heads to $C=0.99$ and evaluate on the contrastive dataset, but only for head-relation pairs that exhibit the highest gradients (top-$k$). This greatly reduces the computational requirements while preserving the robustness of the measurement.

\section{Experiments}

In this section, we investigate the properties of our proposed method of gradient attribution of head-relation pairs. 
We adapt the notation of the relations from \citet{maka-etal-2025-analyzing} but extend the number of relations to include the relation tokens as the input to the network. This is due to the autoregressive nature of LLMs and the fact that the token identity is retained by the Transformer through the layers \citep{brunner2020identifiability}. The relations investigated in this study are as follows (see also Figure~\ref{fig:method} left):
\begin{itemize}[topsep=0pt,itemsep=0pt,partopsep=0pt, parsep=0pt, itemindent=15pt, leftmargin=0pt]
        \item $S_{P} \rightarrow S_{C}$,
        \item $S_{P} \rightarrow S_{C+1}$,
        \item $S_{P+1} \rightarrow S_{C}$,
        \item $S_{P+1} \rightarrow S_{C+1}$,
        \item $T_{P} \rightarrow S_{C}$,
        \item $T_{P} \rightarrow S_{C+1}$,
        \item $T_{P} \rightarrow S_{P}$,
        \item $T_{P} \rightarrow S_{P+1}$,
        \item $T_{P} \rightarrow T_{C}$,
        \item $T_{P} \rightarrow T_{C+1}$,
\end{itemize}
where $S_P$, $T_P$ represent the ambiguous phrases as output in the source and target language respectively, $S_C$, $T_C$ represent the context cue as output (in the source and target language), and the tokens as the input are represented by a $+1$ (e.g., $S_{C+1}$ is the context cue in the source language as the input to the network). 
Note that we do not include the relations from the ambiguous phrase in the target language as input, as at this timestep the model has already decided on the correct (or incorrect) phrase. We use the notation \{layer\}-\{head\} to identify attention heads (e.g., 3-16 represents the 16th head in the third layer of the model).


\begin{table*}[!ht]
\centering
\small
\begin{tabular}{lrrrrrrr}
\hline
\textbf{Loss} & \multicolumn{2}{c}{\textbf{EuroLLM}} & \textbf{EuroLLM} & \multicolumn{3}{c}{\textbf{Qwen}} & \textbf{Gemma}  \\ 
~ & \textbf{Context 5} & \textbf{Context 1} & \textbf{9B*} & \textbf{ContraPro} & \textbf{~~~LCPT} & \textbf{Auxiliary} & ~  \\ 
\hline
$\mathcal{L}_{NLL}$ &  0.0050 &  0.0036 &  0.0008 &  0.0201 &  0.0120 &  0.3562 &  0.0363 \\
\hline
$\mathcal{L}_{logC}$ &  0.3851 &  0.4354 &  0.4624 &  0.4714 &  0.6219 &  0.4832 &  0.2568 \\
$\mathcal{L}_{logTC}$ &  0.4438 &  \textbf{0.5907} &  0.4665 &  0.5440 &  0.6254 &  \textbf{0.5167} &  0.3140 \\
$\mathcal{L}_{TC}$ &  0.4225 &  0.5092 &  0.3704 &  0.5127 &  0.6163 &  0.4970 &  0.3761 \\
\hline
$\mathcal{L}_{logMM}$ &  0.4592 &  0.4883 &  0.5318 &  0.5542 &  0.6180 &  0.4500 &  0.3370 \\
$\mathcal{L}_{logTMM}$ &  0.4846 &  0.5619 &  0.4809 &  0.5692 &  0.6578 &  0.4802 &  0.3448 \\
$\mathcal{L}_{TMM}$ &  \textbf{0.5362} &  0.5709 &  \textbf{0.5606} &  \textbf{0.5838} &  \textbf{0.6799} &  0.4834 &  \textbf{0.4513} \\
\hline
\end{tabular}
\caption{\textbf{Average precision} of finding the head-relation pairs that improve accuracy when modified to $C=0.99$ compared to the unmodified model $\Delta A$ by more than the 99\% confidence interval (of the unmodified model) for all relations of interest for all tested models and phenomena. *EuroLLM 9B was evaluated on the limited set of token-to-token relations.}
\label{tab:average-precision-results}
\end{table*}

We perform a full sweep of Modifying Heads for all relevant relations for the EuroLLM 1.7B instruction-tuned model\footnote{\url{https://huggingface.co/utter-project/EuroLLM-1.7B-Instruct}} \citep{martins2025eurollm} on the ContraPro contrastive dataset \citep{muller-etal-2018-large}. In addition to the current source and target sentences, we concatenate 5 previous sentences on both the source and target side (context size of 5). Additionally, we test the losses on this model with the context size of one. 
Furthermore, we evaluate the methods on the Qwen 2.5 1.5B instruction-tuned model\footnote{\url{https://huggingface.co/Qwen/Qwen2.5-1.5B-Instruct}} \citep{qwen2, qwen2_5} on ContraPro in English-to-German, LCPT in English-to-French \citep{lopes-etal-2020-document}, and Auxiliary phenomenon in English-to-German from the contrastive version of the ctxPro dataset \citep{wicks-post-2023-identifying}. 
Next, we evaluate the third model Gemma 3 1B instruction-tuned \citep{gemma_2025}\footnote{\url{https://huggingface.co/google/gemma-3-1b-it}} on ContraPro.
\ch{The last model we test is the larger EuroLLM 9B Instruct\footnote{\url{https://huggingface.co/utter-project/EuroLLM-9B-Instruct-2512}} \citep{ramos2026eurollm22btechnicalreport} on ContraPro. But, because of the prohibitive computational cost of evaluating each head-relation pair (see Appendix~\ref{sec:time-and-memory-requirements}), we examine the results of the smaller EuroLLM 1.7B model and select 5 relations that achieve the highest increase in accuracy $\Delta A$ ($T_P \rightarrow T_{C+1}$, $T_P \rightarrow T_{P+1}$, $T_P \rightarrow T_{C}$, $T_P \rightarrow S_{C+1}$, $T_P \rightarrow T_{C}$).}
The total number of evaluations are presented in Table~\ref{tab:number-evaluations}, and the time and memory requirements are shown in Tables~\ref{tab:inference-time-memory-requirements} and \ref{tab:gradients-time-memory-requirements} for evaluation and calculating gradients, respectively.

We compare the Token-level Max-Margin loss $\mathcal{L}_{TMM}$ (eq.~\ref{eq:token-max-margin-loss}) with the following losses as baselines:
\begin{itemize}[topsep=0pt,itemsep=0pt,partopsep=0pt, parsep=0pt, itemindent=15pt, leftmargin=0pt]
    \item Negative Log Likelihood of the ambiguous token:
\begin{equation} \label{eq:nll-loss}
\begin{aligned}
\mathcal{L}^d_{NLL} = & -log \Bigl( \bar{P_{\theta}} \bigl( x^{d,c} \bigr) \Bigr),
\end{aligned}
\end{equation}
    \item Contrastive loss for the whole sequence:
\begin{equation} \label{eq:contrastive-loss}
\begin{aligned}
& \mathcal{L}^d_{C} = - \bar{P_\theta} \bigl( X^{d,c} \bigr) + \max_i \Bigl( \bar{P_\theta} \bigl( X^{d,w_i} \bigr) \Bigr),
\end{aligned}
\end{equation}
    \item Contrastive loss for the ambiguous token - similar to \citet{yin-neubig-2022-interpreting},
\begin{equation} \label{eq:token-contrastive-loss}
\begin{aligned}
& \mathcal{L}^d_{TC} = - \bar{P_\theta}\bigl( x^{d,c} \bigr) + \max_i \Bigl( \bar{P_\theta} \bigl(x^{d,w_i} \bigr) \Bigr),
\end{aligned}
\end{equation}
    \item Max-Margin loss for the whole sequence:
\begin{equation} \label{eq:max-margin-loss}
\begin{aligned}
& \mathcal{L}^d_{MM} = \\& \max \biggl(0, \mu - P_\theta \bigl( X^{d,c} \bigr) + \max_i \Bigl( P_\theta \bigl( X^{d,w_i} \bigr) \Bigr) \biggr).
\end{aligned}
\end{equation}
\end{itemize}
Additionally, we include log-space versions of contrastive losses, where probabilities assigned by the model are replaced with their logarithms. We note the losses as $\mathcal{L}_{log X}$ (e.g., $\mathcal{L}_{logTC}$ for the logarithm version of the contrastive loss for the ambiguous token $\mathcal{L}_{TC}$). Because of the issues with numerical stability, we consider only log-probability versions of losses based on the whole sequence ($\mathcal{L}_{logC}$ and $\mathcal{L}_{logMM}$). \ch{We set $\mu=0$, which corresponds to calculating the loss only for the examples where the model was not correct.}

For all tested losses, we aggregate gradients for all head-relation pairs and all examples in the contrastive dataset for which a particular relation is present (context cue inside the context size). We measure Pearson $r$ and Spearman $\rho$ correlation coefficients between the gradients and the accuracy difference $\Delta A$ between the unmodified model and the model with Modified Heads to $C=0.99$ (for the same head-relation pair). \ch{Furthermore, we calculate the average precision scores, Normalized Discounted Cumulative Gain (NDCG), and ranks of retrieving head-relation pairs that result in the accuracy difference $\Delta A$ above the 99\% confidence interval of the unmodified model.} We present the hyperparameters of evaluation in Appendix~\ref{sec:hypoerparameters}.

\subsection{Gradient-Modifying Alignment}
\label{sec:gradient-alignment}

\begin{figure*}
    \centering
    \includegraphics[width=1\linewidth]{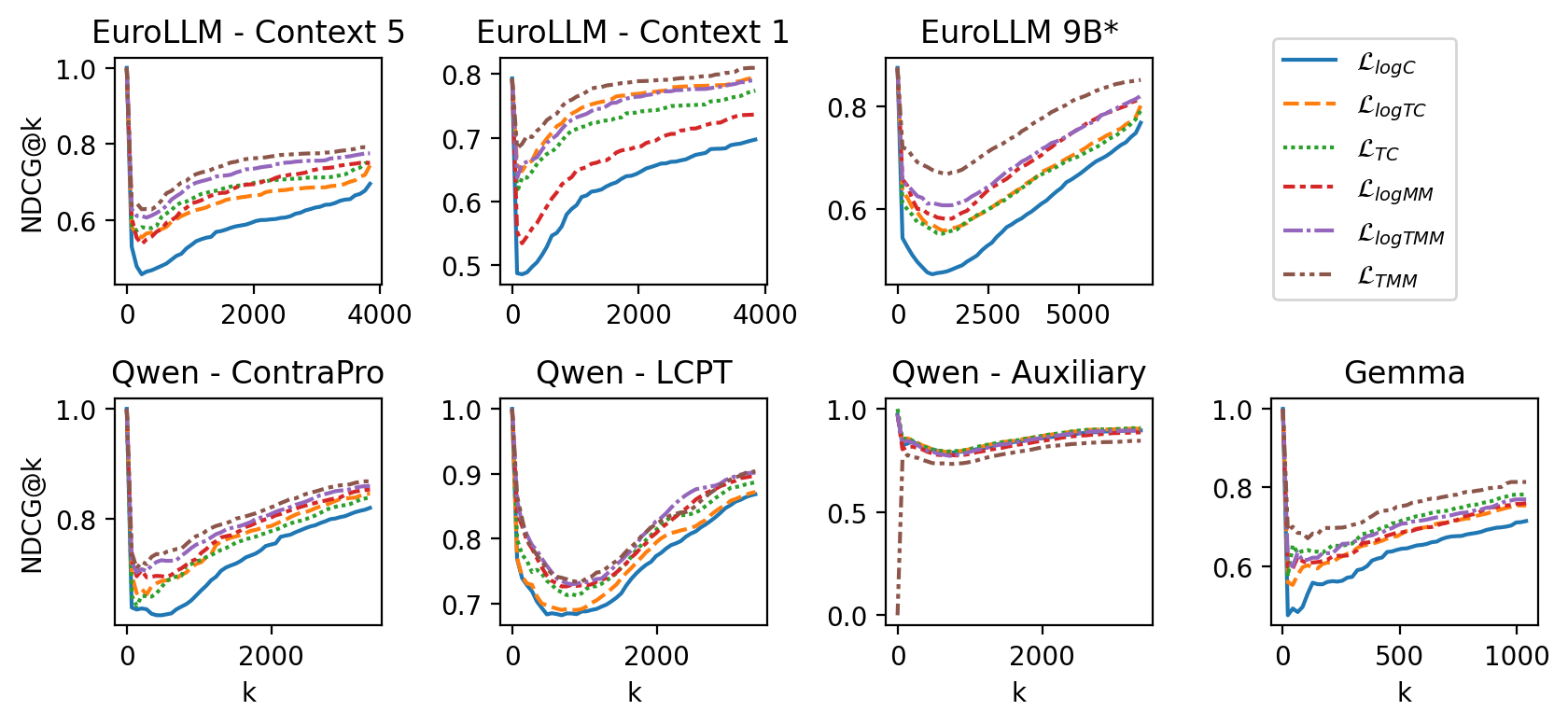}
    \caption{The \textbf{Normalized Discounted Cumulative Gain (NDCG)} of finding the head-relation pairs that improve accuracy when modified to $C=0.99$ compared to the unmodified model $\Delta A$ by more than the 99\% confidence interval (of the unmodified model) for tested losses. *EuroLLM 9B was evaluated on the limited set of token-to-token relations.}
    \label{fig:eurollm-ndcg}
\end{figure*}


We present the baseline results for all models and phenomena in Table~\ref{tab:all-base-results} in Appendix~\ref{sec:extended-results}.
The results in terms of Pearson $r$ and Spearman $\rho$ correlation coefficients for the tested models can be seen in Table~\ref{tab:full-correlation-results} in Appendix~\ref{sec:extended-results}. We additionally include the correlations of two subsets of head-relation pairs: those with an accuracy difference greater than zero ($\Delta A > 0$) and the remaining ones ($\Delta A \le 0$). This is motivated by the fact that fundamentally different mechanisms are responsible for both effects. While the model's positive response to Modifying Heads to $C=0.99$ can be interpreted precisely (the attention score for a particular head and relation is increased), the negative response can be attributed to a reduction in the attention score for another token that is important for the head. 
This can be observed in Figure~\ref{fig:gradient-accuracy-comparison} in Appendix~\ref{sec:extended-results}, where we show the comparison between the aggregated gradients and the accuracy difference $\Delta A$ for the EuroLLM 1.7B Instruct model.
Token-level Max-Margin loss $\mathcal{L}_{TMM}$ shows the highest Spearman correlations $\rho$ for accuracy differences greater than zero ($\Delta A > 0$) for most of the tested scenarios. The exceptions were the Qwen model on Auxiliary and LCPT datasets. 

\begin{figure*}
    \centering
    \includegraphics[width=1\linewidth]{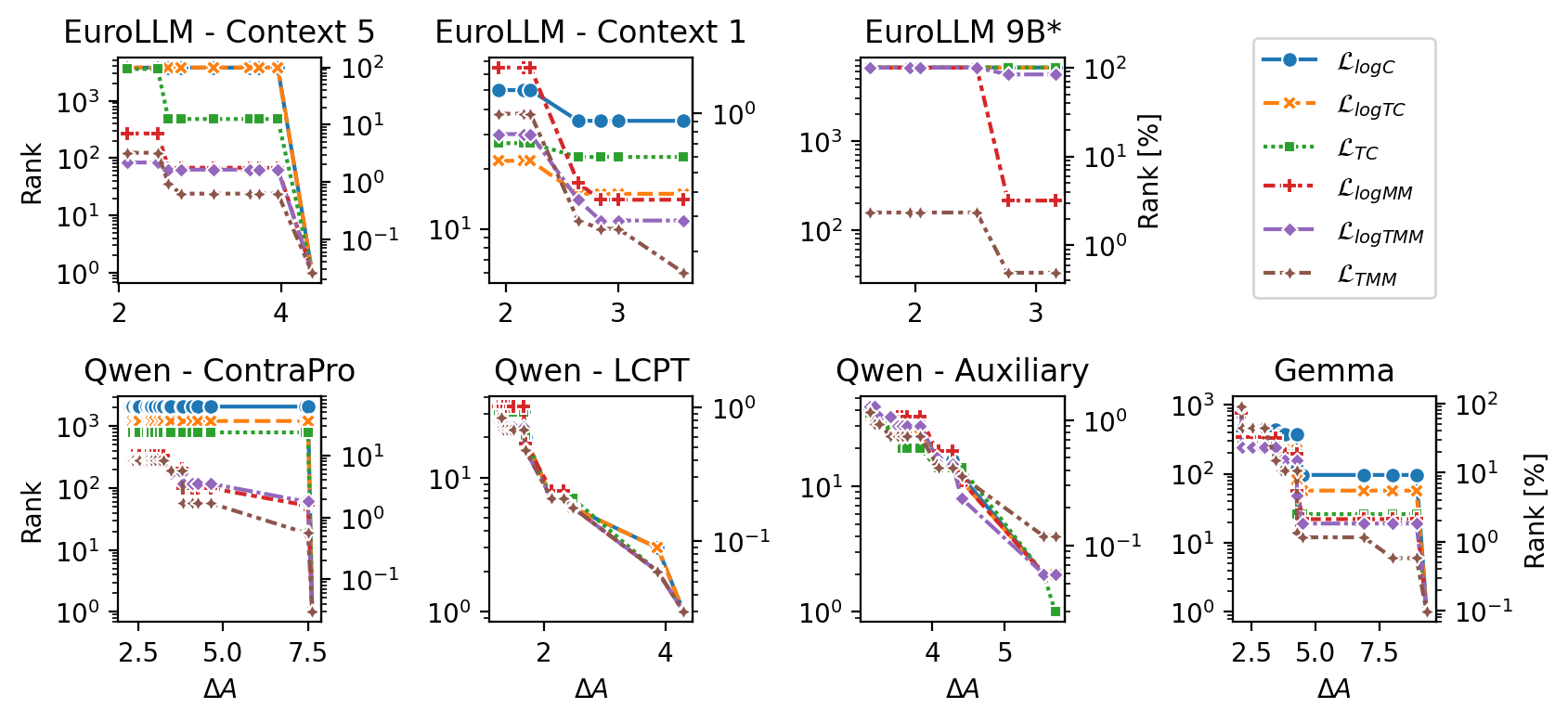}
    \caption{The \textbf{ranks} of head-relation pairs according to the gradients based on the tested losses for all tested models and phenomena. For clarity, we show the head-relation pairs that improve accuracy compared to the unmodified model $\Delta A$ by more than the 99\% confidence interval (of the unmodified model). *EuroLLM 9B was evaluated on the limited set of token-to-token relations.}
    \label{fig:eurollm-ranks-top}
\end{figure*}

\ch{Table~\ref{tab:average-precision-results} shows the results in terms of the average precision of retrieving the head-relation pairs that improve the accuracy $\Delta A$ by more than 99\% confidence interval of the unmodified model. Token-level Max-Margin loss $\mathcal{L}_{TMM}$ achieves the best results in most of the cases, with the exceptions of EuroLLM 1.7B with context size of 1 (where it achieved the second best score), and Qwen on the Auxiliary task. In Figure~\ref{fig:eurollm-ndcg} we show the Normalized Discounted Cumulative Gain (NDCG) for all tested methods, where the proposed loss achieved the best results for most scenarios. The exceptions are Qwen on LCPT and Auxiliary datasets.
Finally, we examined the ranks of head-relation pairs in terms of the accuracy difference $\Delta A$ given by the backpropagated losses. The results can be seen in Figure~\ref{fig:eurollm-ranks-top}. Ranking according to the Token-level Max-Margin loss $\mathcal{L}_{TMM}$ outperforms other losses for most of the tested models and phenomena.}

\ch{While Token-level Max-Margin loss $\mathcal{L}_{TMM}$ outperforms other losses in most tested scenarios, it was surpassed on EuroLLM 1.7B with context size of 1 and Qwen on LCPT and Auxiliary datasets. We argue for using it for all models and phenomena, because it is still able to retrieve all relevant head-relation pairs in those scenarios within the top 1\% of the ranking (see Figure~\ref{fig:eurollm-ranks-top}).}

\subsection{Why Gradients Work?}
\label{sec:why-gradients-work}

\begin{figure*}[!ht]
\center{}
\hfill
    \begin{subfigure}{0.4\linewidth}
        \includegraphics[width=1\linewidth]{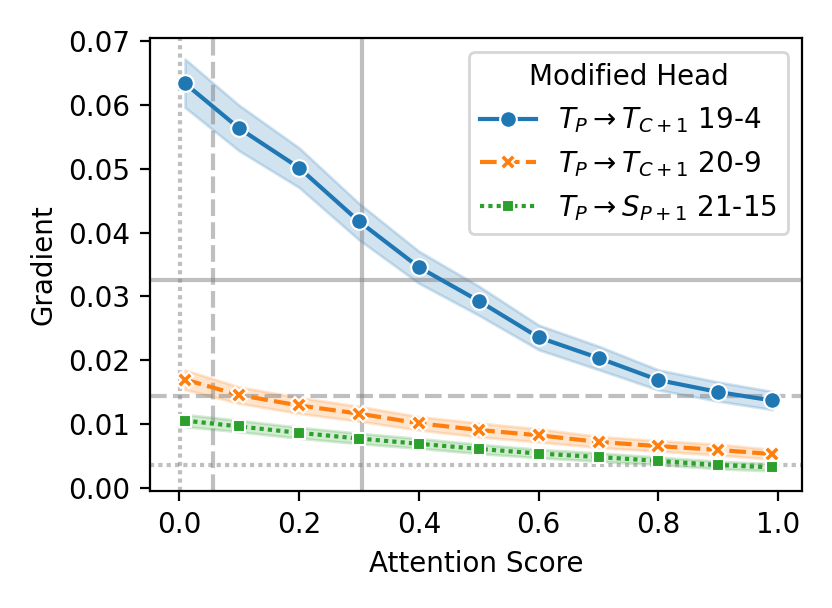}
        \caption{Aggregated}
        \label{fig:modified-gradients-aggregated}
    \end{subfigure}
    \hfill
    \begin{subfigure}{0.4\linewidth}
        \includegraphics[width=1\linewidth]{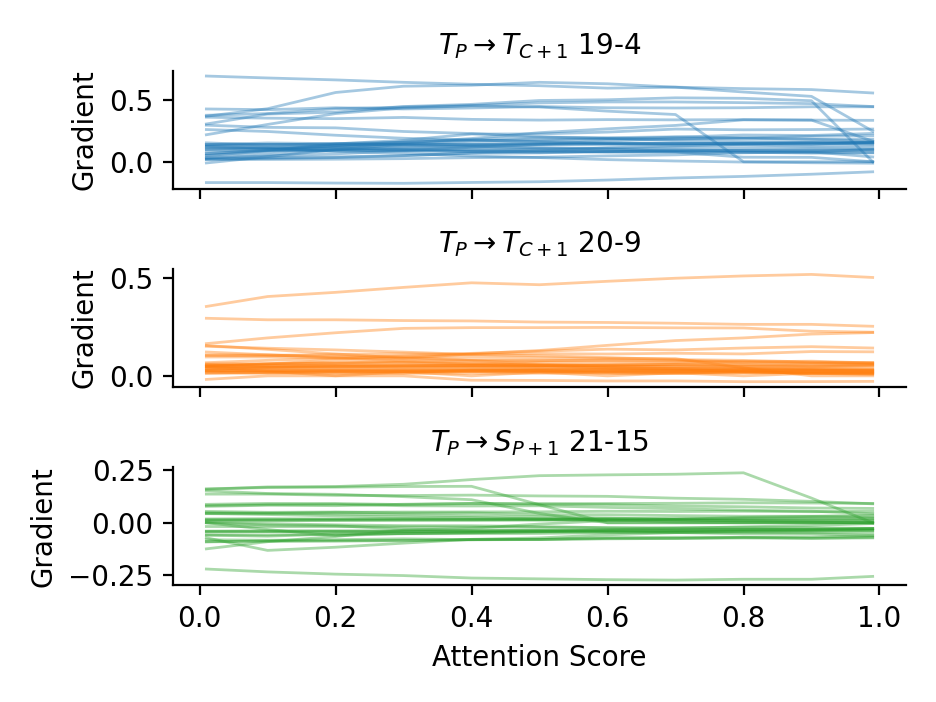}
        \caption{Selected Examples}
        \label{fig:modified-gradients-examples}
    \end{subfigure}
    \hfill\hfill
    \caption{The gradients (based on the Token-level Max-Margin Loss $\mathcal{L}_{TMM}$) of selected heads when modified to different values of Attention Scores $C$ when: aggregated (\ref{fig:modified-gradients-aggregated}), and sampled 30 examples with non-zero gradient for most head-relation pairs and $C$ values (\ref{fig:modified-gradients-examples}).}
    \label{fig:modified-gradients}
\end{figure*}

While gradient-attribution methods have seen successes \citep{10.5555/3305890.3306024, yin-neubig-2022-interpreting, song-etal-2024-better, 11084559}, they have been criticized for their unreliability as explanations \citep{wang-etal-2020-gradient}. In this section, we try to answer the question of why our method with Token-level Max-Margin loss $\mathcal{L}_{TMM}$ showed moderate to strong correlations with the accuracy difference $\Delta A$ and effectively ranked top head-relation pairs. 

\begin{figure*}
    \centering
    \includegraphics[width=0.99\linewidth]{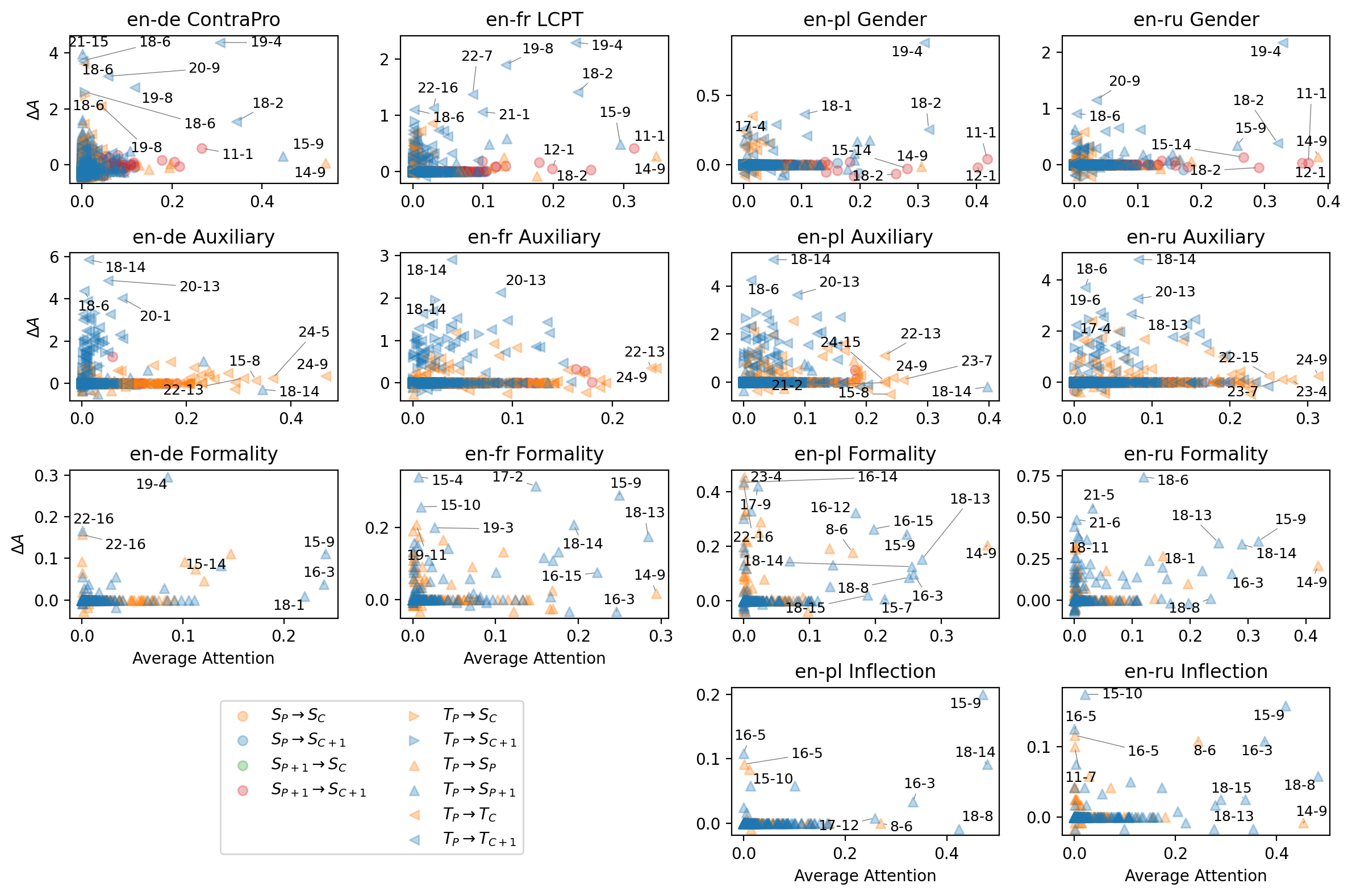}
    \caption{The results in terms of average attention scores (x axes) and accuracy increase $\Delta A$ (y axes) for head-relation pairs for \textbf{EuroLLM 1.7B Instruct} on all tested phenomena and language directions. To improve clarity, we omitted the head-relation pairs with a large negative $\Delta A$.}
    \label{fig:eurollm-attention-accuracy}
\end{figure*}

We calculate gradients for three selected heads when modified to values in $C \in \{0.01, 0.1, 0.2, ..., 0.9, 0.99\}$. Doing this allows us to see the monotonicity of the gradient-accuracy increase relation. While the relation for the accumulated gradients (shown in Figure~\ref{fig:modified-gradients-aggregated}) is monotonic (to the granularity of the tested values), the gradients for the individual examples (depicted in Figure~\ref{fig:modified-gradients-examples}) are not only non-monotonic but also change sign. This finding highlights that gradient-attribution methods are hindered by the non-monotonicity of the underlying function, and the accumulation of the gradients can mitigate this issue \citep{10.5555/3305890.3306024}.

\section{Analysis}

In this section, we analyze the models on a wide range of language directions and phenomena. Specifically, we include translating from English to German, French, Polish, and Russian, and the following phenomena: Gender (we use ContraPro and LCPT for English-to-German and English-to-French, respectively, and ctxPro Gender for English-to-Polish and English-to-Russian), Auxiliary, Formality, and Inflection (the last one available only for English-to-Polish and English-to-Russian). \ch{This amounts to 50 scenarios across four models and four language directions.}

We apply our gradient-based framework by first backpropagating the Token-level Max-Margin loss $\mathcal{L}_{TMM}$ and selecting top-$k$ head-relation pairs in terms of aggregated gradient. Motivated by the rank of top head-relation pairs (see Figure~\ref{fig:eurollm-ranks-top}), we set the number of heads evaluated by Modifying Heads to be equal to 3\% of all head-relation pairs. Note that, due to the lack of specified context cues in the ctxPro dataset for Formality and Inflection, we cannot define all of the relations explored previously for those phenomena, as only the contextually dependent phrases are identified. This shortens the list of relations to: $T_P \rightarrow S_P$ and $T_P \rightarrow S_{P+1}$, effectively reducing the number of head-relation pairs. For those phenomena, we still maintain the same computational budget for modifying top heads as for other phenomena. Therefore, we modify 15\% of top head-relation pairs for Formality and Inflection. While gradients align well with the change in the accuracy $\Delta A$, we saw that some heads can attend a relation but the model will not improve the performance when those heads are modified (e.g., 14-9 $T_P \rightarrow S_P$ in Figure~\ref{fig:eurollm-attention-accuracy}). To analyze those heads, we modify 0.3\% of head-relation pairs with the highest average attention scores. \ch{Because the evaluation of the EuroLLM 9B model on ContraPro was done for a limited set of relations, we additionally evaluate the top 3\% head-relation pairs according to our method and analyze the combined results. We omit evaluating Formality and Inflection for this model due to the low statistical significance of the results for those phenomena on the smaller models.}

In general, by focusing our computational resources exclusively on the promising head-relation pairs, we were able to scale the analysis to 50 phenomena across 4 different language directions and 4 models in total. This allowed us to assess the cross-lingual and cross-phenomena behavior of the models while reducing the computational cost of the analysis by more than 96\%. The results can be found in Figures~\ref{fig:eurollm-attention-accuracy} (EuroLLM 1.7B Instruct), \ref{fig:eurollm-9b-attention-accuracy} (EuroLLM 9B Instruct), \ref{fig:qwen-attention-accuracy} (Qwen 2.5 1.5B Instruct), and \ref{fig:gemma-attention-accuracy} (Gemma 3 1B Instruct).

\subsection{Results}

Our work confirms that heads appear to have specific functions by attending to a specific relation or improving the model's performance when modified to $C=0.99$. While this is in line with previous research \citep{clark-etal-2019-bert, voita-etal-2019-analyzing, Olsson2022IncontextLA, maka-etal-2025-analyzing}, our analysis reveals that the heads can attend to \ch{or} benefit from modifying multiple relations. This generalization of capability can manifest in several forms spanning different types of relations:
\begin{itemize}[topsep=0pt,itemsep=0pt,partopsep=0pt, parsep=0pt, itemindent=15pt, leftmargin=0pt]
    \item Same relation but the target token on the input and output - for example, the heads 19-8 in EuroLLM 1.7B and 24-11 in Qwen 2.5 1.5B improve accuracy on the ContraPro dataset when modified to attend both the $T_P \rightarrow T_C$ and $T_P \rightarrow T_{C+1}$ relations. This corroborates the finding that the identity of the input token is maintained through the layers \citep{brunner2020identifiability}, but such heads are not widespread.
    \item Same relation and phenomenon but different language directions - the examples include head-relation pairs 14-9 $T_P \rightarrow S_P$ in EuroLLM 1.7B, 32-3 $T_P \rightarrow T_{C+1}$ in EuroLLM 9B, 16-2 $T_P \rightarrow T_{C+1}$ in Gemma 3 1B for ContraPro, LCPT, and Gender (both for English-to-Polish and English-to-Russian), and 20-13 $T_P \rightarrow T_{C+1}$ and 22-13 $T_P \rightarrow T_C$ for Auxiliary (all language directions) in EuroLLM 1.7B, among others.
    \item Same relation but different language direction and phenomenon - for example, head-relation pairs 16-3 $T_P \rightarrow S_{P+1}$ for Formality and Inflection \ch{(high attention but not accuracy difference)} in EuroLLM 1.7B, and 12-2 $T_P \rightarrow S_{P+1}$ in Gemma 3 1B exhibit this behavior.
    \item Different relation, language direction, and phenomenon - most notably, heads 18-6, 
    and 18-14 in EuroLLM 1.7B, 25-4 in EuroLLM 9B, 20-2 in Qwen 2.5 1.5B, and 18-3 in Gemma 3 1B.
\end{itemize}
The existence of those "general-purpose" attention heads suggests that the updates to the residual stream coming from those heads share some component, as the information they encode can improve the model's performance on several tasks. We leave the detailed exploration of this finding for future work. 

Importantly, the results for Formality and Inflection in terms of the accuracy difference $\Delta A$ when modified to $C=0.99$ are not statistically significant (see Table~\ref{tab:all-base-results} for the confidence intervals of the results of the unmodified models). We hypothesize that this can be caused by the limited set of relations we investigated (due to the lack of context cues being annotated by the ctxPro dataset). Additionally, modifying to $C=0.99$ requires the model not to achieve a perfect score, as there needs to be room for improvement to observe any positive change in accuracy. This shows a limitation of the Modifying Heads technique: the need for well-defined relations between tokens and evaluation metrics that are challenging for the models.

\begin{figure}
    \centering
    \includegraphics[width=0.9\linewidth]{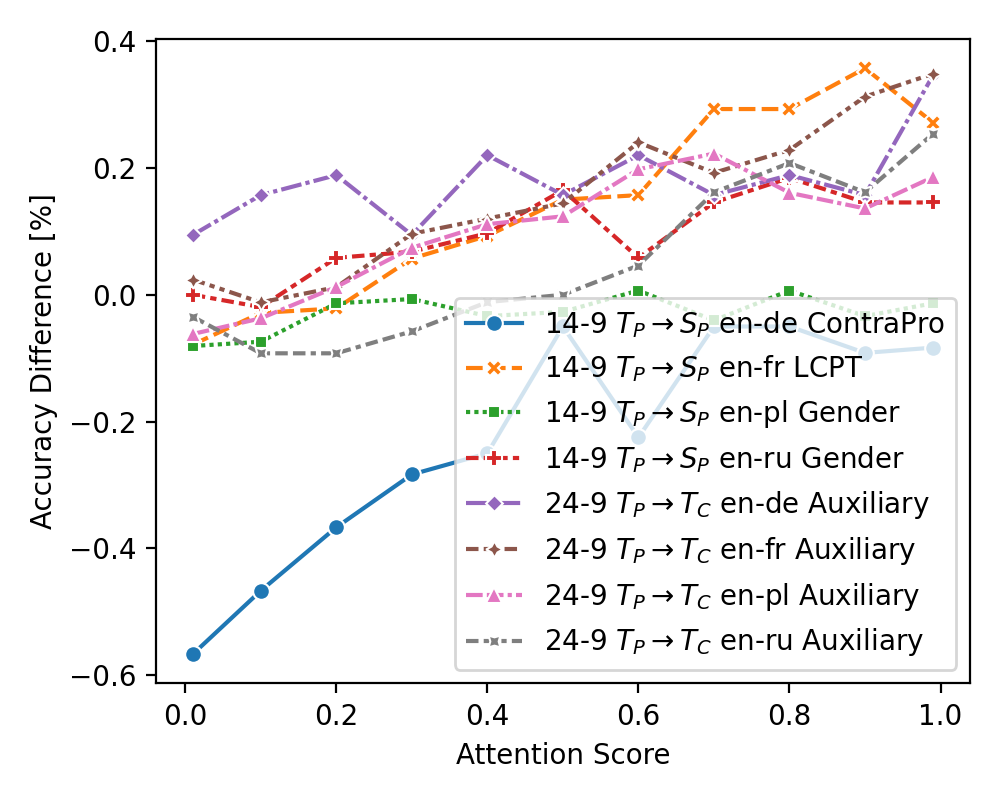}
    \caption{Accuracy difference $\Delta A$ of selected head-relation pairs for different values of modified attention scores $C$ for EuroLLM 1.7B Instruct.}
    \label{fig:eurollm-detailed_modified}
\end{figure}

\begin{figure*}[!ht]
\center{}
    \begin{subfigure}{0.45\linewidth}
        \includegraphics[width=1\linewidth]{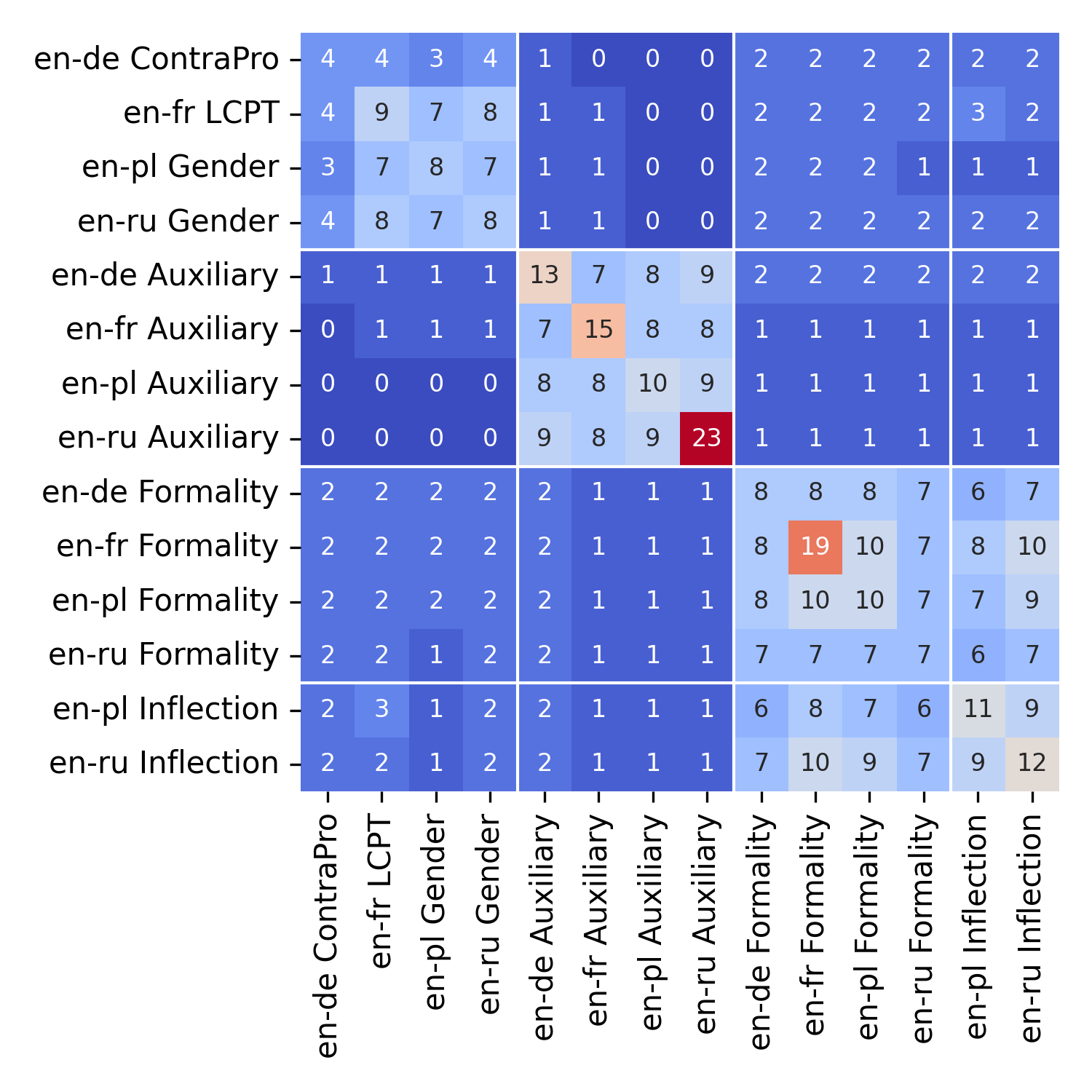}
        \caption{Average Attention}
        \label{fig:eurollm-heads-overlap-attention}
    \end{subfigure}
    %
    \begin{subfigure}{0.45\linewidth}
        \includegraphics[width=1\linewidth]{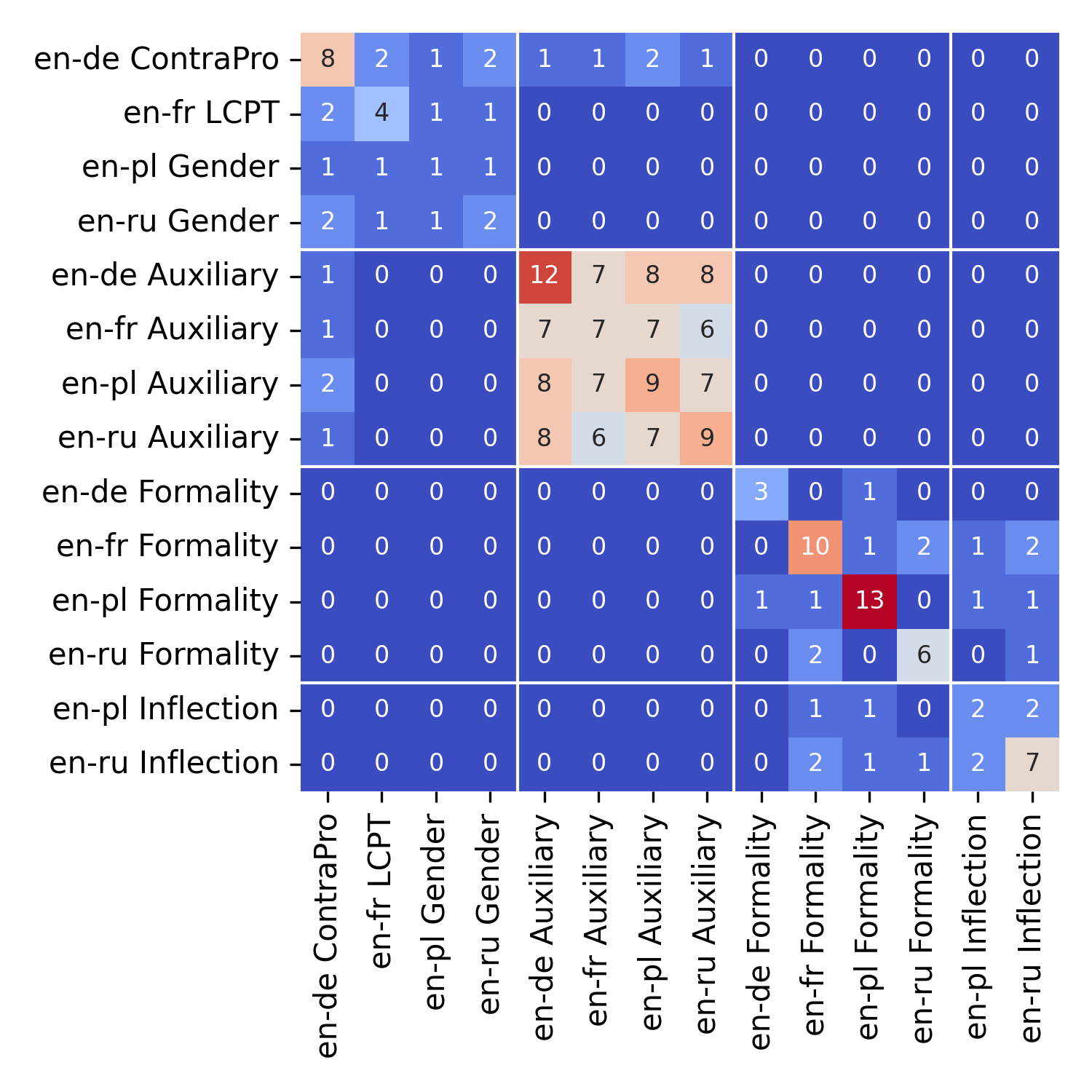}
        \caption{$\Delta A$}
        \label{fig:eurollm-heads-overlap-modified}
    \end{subfigure}\hfill
    \caption{Number of shared head-relation pairs between language directions and phenomena for EuroLLM 1.7B Instruct. We select heads that exhibit average attention (\ref{fig:eurollm-heads-overlap-attention}) and $\Delta A$ (\ref{fig:eurollm-heads-overlap-modified}) higher than half of the maximum value for a specified language direction and phenomenon.}
    \label{fig:eurollm-heads-overlap}
\end{figure*}

\subsection{Unresponsive Heads}

Many heads show high average attention scores with no improvement in accuracy when modified to $C=0.99$ for specific relations. In EuroLLM 1.7B, those include head-relation pairs 14-9 $T_P \rightarrow S_P$ for Gender and Formality, 11-1 $S_{P+1} \rightarrow S_{C+1}$ for Gender, 24-9 $T_P \rightarrow T_C$ for Auxiliary, and 18-8 $T_P \rightarrow S_{P+1}$ for Inflection (among others). We investigate whether this is due to those heads being "optimal" in their apparent function of attending to the relations. To test this hypothesis, we modify selected heads (namely, 14-9 $T_P \rightarrow S_P$ for Gender and 24-9 $T_P \rightarrow T_C$ for Auxiliary in all language directions) to the values of $C$ spanning the whole range: $C \in \{0.01, 0.1, 0.2, ..., 0.9, 0.99\}$. If our hypothesis is true, we expect to see a plateau for $C$s larger than the average attention score of the unmodified model. 

The results can be seen in Figure~\ref{fig:eurollm-detailed_modified}. While we do not see a statistically significant improvement above the average attention scores, we also do not observe the drop in accuracy $\Delta A$ below that point (14-9 $T_P \rightarrow S_P$ for ContraPro shows slightly larger drop in accuracy, but is still within the confidence interval of the unmodified model; see Table~\ref{tab:all-base-results}). This suggests that even though the heads in question attend to the relations, they are not crucial to the model for disambiguating the context-dependent phrases. This corroborates the notion that raw attention scores alone are not enough to interpret the models \citep{jain-wallace-2019-attention}. We hypothesize that the existing redundancies inside the model (other heads attending the same relation, such as 15-9 $T_P \rightarrow S_{P+1}$ for Gender) enable the model to function properly.

\subsection{Link to Generalizability}

Motivated by the overlap of some heads between phenomena and language directions, we quantified the extent of shared head-relation pairs between each phenomena. We selected the head-relation pairs that exhibit an increase in accuracy $\Delta A$ greater than half of the maximum $\Delta A$ for each phenomenon and language direction separately. We applied the same criterion to average attention scores. The results can be seen in Figure~\ref{fig:eurollm-heads-overlap}. 

We observed a large overlap for Auxiliary between language directions for both the average attention scores and accuracy increases $\Delta A$. For Gender (including ContraPro and LCPT), the overlap is high based on the average attention, and substantially lower based on the accuracy increase $\Delta A$. Similar trends can be seen for Formality and Inflection, although the results in terms of accuracy increase $\Delta A$ are not statistically significant. We argue that the overlap based on the accuracy increase $\Delta A$ better explains the transfer during training between phenomena observed by \citet{maka-etal-2025-train}, where it was higher for Auxiliary, very limited inside other phenomena, and not present at all between different phenomena. Intuitively, positive changes to accuracy when modified indicate the potential of certain heads to improve the model's performance during training. Whether these heads actually learn to attend more to the relations during training and how this relates to the improvements in other phenomena is left for future work.

\subsection{Tuning Attention Heads}

\begin{table}[!ht]
\centering
\tiny
\begin{tabular}{llcccccc}
\hline
\multicolumn{2}{l}{\textbf{Model}} & \textbf{BLEU} & \textbf{COMET} & \multicolumn{2}{c}{\textbf{ContraPro}} \\
\hline
\multicolumn{2}{l}{Baseline} & 21.41 & 0.7862 & \multicolumn{2}{c}{83.14\%} \\ 
\hline
\textbf{Head} & \textbf{Relation} & \textbf{BLEU} & \textbf{COMET} & \textbf{Modified} & \textbf{Tuned} \\
\hline
19-4 & $T_{P} \rightarrow T_{C+1}$ & 21.33 & 0.7857 & +2.84\% & +2.72\% \\
18-6 & $T_{P} \rightarrow T_{C+1}$ & 21.19 & 0.7862 & +3.00\% & +2.25\% \\
18-6 & $T_{P} \rightarrow T_{C}$ & 20.90 & 0.7858 & +2.64\% & +1.95\% \\
21-15 & $T_{P} \rightarrow S_{P+1}$ & 21.39 & 0.7861 & +3.58\% & +1.00\% \\
\hline
\end{tabular}
\caption{BLEU \& COMET scores on English-to-German OpenSubtitles of the EuroLLM 1.7B model with context size of 1 and the accuracy on the ContraPro dataset compared between modified and tuned head-relation pairs.}
\label{tab:head-tuning-results}
\end{table}


\ch{The Modifying Heads technique artificially intervenes in the model's forward pass. In this section, we examine how this intervention, when learned by the model, impacts the general translation performance and the contrastive accuracy.
To this end, we replicate the Head Tuning technique from \citet{maka-etal-2025-analyzing}, where the parameters of $Q$ and $K$ projections of the selected heads are tuned to attend more to the specified relation. We tune the heads of the EuroLLM 1.7B model with a context size of 1 using the ctxPro English-to-German Gender dataset. Next, we measure ContraPro accuracy and BLEU and COMET on the 10000 randomly selected English-to-German examples from the OpenSubtitles 2018 \citep{lison-etal-2018-opensubtitles2018} dataset. The details are presented in Appendix~\ref{sec:head-tuning-details}.}

\ch{The results for the four selected head-relation pairs are presented in Table~\ref{tab:head-tuning-results}. The BLEU and COMET scores are stable compared to the base model, showing that when the change in attention scores is localized, the effect on general translation quality is minimal. The improvement in ContraPro accuracy can vary between head-relation pairs. While tuning the 19-4 $T_{P} \rightarrow T_{C+1}$ head-relation pair almost replicates the improvement of modifying, head tuning 21-15 $T_{P} \rightarrow S_{P+1}$ leads to an improvement of 1 percentage point compared to more than 3.5 when modified.}

\section{Indirect Object Identification Task}

\ch{To test the generalizability of our method, we apply it to the task of Indirect Object Identification (IOI) \citep{wang2023interpretability}, where an expression such as "Then, [B] and [A] went to the house. [B] gave a drink to" (where "[A]" and "[B]" are names) should be completed with "[A]". We used EuroLLM 1.7B Instruct, but the task is very simple, and even this small model achieves near-perfect accuracy of 99.9\%. Therefore, we follow the original paper and measure the difference in logits between the correct and incorrect name. The tested model exhibits a logit difference of $5.26$. The details are presented in Appendix~\ref{sec:ioi-details}.}

\begin{figure}
    \centering
    \includegraphics[width=1\linewidth]{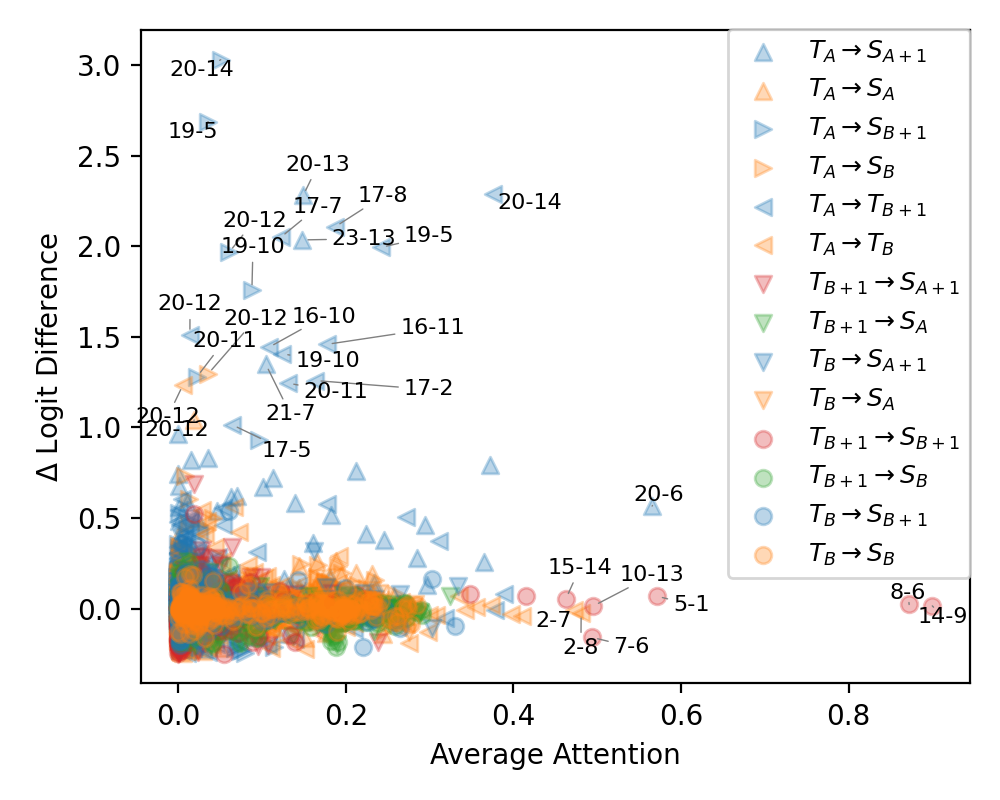}
    \caption{Average attention scores (x axis) and the change in the logit difference when modified to $C=0.99$ for each head and relation on IOI dataset for EuroLLM 1.7B Instruct. For readability, we only show head-relation pairs with the positive change in logit difference.}
    \label{fig:eurollm-ioi}
\end{figure}

\ch{
We consider the token-to-token relations between the two names appearing in the examples. We mark the names in the first part of the IOI sequence as $S_A$ and $S_B$ for names "[A]" and "[B]", respectively, and as $T_A$ and $T_B$ in the second part.
The results of modifying heads are presented in Figure~\ref{fig:eurollm-ioi}, where the greatest improvement of more than 57\% in the metric is achieved by 20-14 $T_A \rightarrow S_{B+1}$. Similar to the Context-aware MT setting, head-relation pairs that already show high average attention do not, in general, increase the metric when modified. Furthermore, we found heads that increase the metric when modified to attend different relations. For example, head 20-12 responds to $T_A \rightarrow S_{B+1}$, $T_A \rightarrow S_{B}$, $T_A \rightarrow T_{B+1}$, $T_A \rightarrow T_{B}$, and $T_A \rightarrow S_{A}$.}

\ch{To test our method, we backpropagate the loss to the attention maps. The high accuracy on this task motivates the use of the log-probability Contrastive loss $\mathcal{L}_{logTC}$ (instead of the Max-Margin loss), which is directly related to the metric of the difference in the logits between "[A]" and "[B]". Using this loss with our method achieves Pearson and Spearman correlations of $0.6700$ and $0.6862$, respectively, and average precision of $0.6505$. Furthermore, the top 3\% head-relation pairs contain all results with a change in logit difference greater than $0.60$ (top 37 head-relation pairs), while reducing the computational cost by almost 97\%. This supports the generalizability and efficacy of our method.}

\section{Conclusions}

In this work, we propose a gradient-based head attribution method based on the contrastive loss backpropagated to the attention maps. We exhaustively evaluate the method on the established tasks of disambiguating context-dependent phrases in Context-aware Machine Translation. We show that our method correlates with the Modifying Heads technique, allowing us a substantial expansion of the analysis scope, by reducing the computational cost by more than 96\%. In total, we analyze 50 phenomena across 4 models and 4 language directions. \ch{We demonstrate the potential generalizability of our method by applying it to the task of Indirect Object Identification.}

We identify that some heads are "general-purpose", improving the model's performance when attending to different relations. This behavior generalizes across different language directions and in some cases phenomena, and we link this overlap in the heads' apparent functions to the transfer effect between language directions and phenomena during training. Finally, our findings suggest that the models develop redundancies during training in terms of the head functions. In future work, we will expand the analysis to other contextual tasks and larger models.




\bibliography{custom,anthology-1,anthology-2}
\bibliographystyle{acl_natbib}

\onecolumn

\appendix

\section{Generating Contrastive Dataset}
\label{sec:contrastive-ctxpro}

In this section, we describe the creation process of the contrastive version of the ctxPro datasets \citep{wicks-post-2023-identifying}. 
Below, we list the contextually-dependent linguistic phenomena that we included in our study:
\begin{itemize}[topsep=3pt,itemsep=3pt,partopsep=0pt, parsep=0pt, itemindent=15pt, leftmargin=0pt]
    \item \textbf{Gender} (anaphoric pronouns) - translating a pronoun from a non-gendered language to a language with gendered nouns.

    \item \textbf{Formality} (anaphoric pronouns) - translating into a language with different second-person pronouns distinguishing intimate from formal relationships between speakers from a language lacking this distinction.
    

    \item \textbf{Auxiliary} (verb phrase ellipsis) - translating into a language that require the head of the verb phrase from a language that allows for only the modal or auxiliary to be used.
    
    \item \textbf{Inflection} (verb phrase ellipsis) - translating into a language with noun morphology dependent on the grammatical role from a language where this is not the case.
\end{itemize}

We started with the evaluation datasets released with the paper under \url{https://github.com/rewicks/ctxpro/tree/main/paper/jsons}. For all phenomena and language directions, we replaced the contextually dependent phrase in the target segment with phrases sampled from the list of phrases from the other examples (excluding the same phrases), respecting the capitalization of the original phrase. We sampled with respect to the distribution of occurrences, as our tests showed that this increased the difficulty of the created datasets. We chose the number of contrastive target sentences for each example to balance the difficulty (based on the EuroLLM 1.7B model) and the computational cost of evaluation. We set this number to 4 for gender, 5 for formality and inflection, and 8 for auxiliary. In the experiments, we used ContraPro \citep{muller-etal-2018-large} and LCPT \citep{lopes-etal-2020-document} instead of the generated contrastive versions of ctxPro for Gender in English-to-German and English-to-French, respectively.

\section{Evaluation Details}
\label{sec:hypoerparameters}

We evaluated the models on a single GPU (NVIDIA GeForce RTX 3090 24GB for EuroLLM 1.7B Context 1, Qwen 2.5 1.5B, and Gemma 3 1B, and NVIDIA H100 80GB for EuroLLM 1.7B Context 5 and EuroLLM 9B). We used 5 previous sentences as context on both the source and target side (apart from the experiment with EuroLLM 1.7B Context 1). We used a batch size of 2 for evaluation and 1 for backpropagating gradients. The models were loaded with precision bf16 (all models from the EuroLLM family) and 8-bit (Qwen 2.5 and Gemma 3). We set the maximum length of the prompt to 2048 for EuroLLM 1.7B and Gemma 3 1B, and 4096 for EuroLLM 9B and Qwen 2.5 1.5B.

To prepare the inputs to the models, we concatenated the context and current sentences on both the source and target sides. We tokenized the prompt shown in Listing~\ref{lst:eurollm-1-7b-prompt} using the models' tokenizers and passed it through the models to obtain the token probabilities. We used the system prompts found in the model cards on HuggingFace. The sizes of the used datasets are presented in Table~\ref{tab:dataset-sizes}.

\begin{table}[!ht]
\centering
\small
\begin{tabular}{llrr}
\hline
\textbf{Language Direction} & \textbf{Phenomenon} & \textbf{Examples} & \textbf{Contrastive} \\
\hline
English-to-German & ContraPro & 12000 & 3.0 \\
~ & Gender & 13640 & 5.0 \\
~ & Auxiliary & 3180 & 9.0 \\
~ & Formality & 10800 & 6.0 \\
\hline
English-to-French & LCPT & 14000 & 2.0 \\
~ & Gender & 19098 & 5.0 \\
~ & Auxiliary & 8322 & 9.0 \\
~ & Formality & 12000 & 5.0 \\
\hline
English-to-Polish & Gender & 14855 & 5.0 \\
~ & Auxiliary & 8085 & 9.0 \\
~ & Formality & 15216 & 6.0 \\
~ & Inflection & 12000 & 6.0 \\
\hline
English-to-Russian & Gender & 10305 & 5.0 \\
~ & Auxiliary & 8667 & 9.0 \\
~ & Formality & 10075 & 6.0 \\
~ & Inflection & 12000 & 6.0 \\
\hline
\end{tabular}
\caption{The sizes of the used datasets in terms of the number of examples and the average number of contrastive targets (including the correct target) for each dataset and language direction.}
\label{tab:dataset-sizes}
\end{table}

\begin{code}[caption={Input template used for evaluating the models. The concatenated context and current sentences on the source and target sides are represented as \texttt{\{src\}} and \texttt{\{tgt\}} respectively. The source and target languages are specified in natural language (e.g., "English", "German") and marked as \texttt{\{src\_lang\}} and \texttt{\{tgt\_lang\}} respectively. We used the system prompts found in the model cards on HuggingFace.}, label={lst:eurollm-1-7b-prompt}]
{
  "role": "system",
  "content": {system_prompt}
},
{
  "role": "user",
  "content": "Translate the following {src_lang} text to {tgt_lang}: '{src}'"
},
{
  "role": "assistant",
  "content": {tgt}
}
\end{code}

\section{Time and Memory Requirements}
\label{sec:time-and-memory-requirements}

The Modifying Heads technique requires repeated evaluations for intervening in attention for each head-relation pair. Table~\ref{tab:number-evaluations} shows the number of layers, heads, and evaluations (assuming 10 token-to-token relations that we used in this study) for each tested model. The largest of our models (EuroLLM 9B Instruct) requires 133400 separate evaluations.

In Tables~\ref{tab:inference-time-memory-requirements} and \ref{tab:gradients-time-memory-requirements}, we present the average execution time and peak GPU memory\footnote{We used the \texttt{torch.cuda.memory.max\_memory\_allocated()} function.} for every model, language direction, and phenomenon. The calculation of gradients used more GPU memory compared to the evaluation, even with the lower batch size. Although the execution time is also increased, this increase is small in comparison to the total time required for the full sweep over the head-relation pairs, because of the very high number of evaluations. For example, assuming evaluating 3\% of the top head-relation pairs, the total computation time for the EuroLLM 9B Instruct model on ContraPro is 175 GPU hours. Meanwhile, the full sweep would take 5824 GPU hours (our method requires 3\% of the full sweep time). The proportion of time spent on calculating gradients is smaller for models containing more heads. For Gemma 3 1B with only 104 heads in the model, the full sweep of ContraPro requires 537 GPU hours compared to 21 hours for our method (4\% of the full sweep).

\begin{table*}[!ht]
\centering
\begin{tabular}{lrrrrr}
\hline 
\textbf{Model} & \textbf{Layers} & \textbf{Heads} & \textbf{Heads Total} & \textbf{Evaluations} & \textbf{Our Method} \\ 
\hline
 EuroLLM 1.7B Instruct & 24 & 16 & 384 & 3840 & 120 \\
 EuroLLM 9B Instruct & 42 & 32 & 1344 & 13440 & 400 \\ 
 Qwen 2.5 1.5B Instruct & 28 & 12 & 336 & 3360 & 100 \\ 
 Gemma 3 1B Instruct & 24 & 4 & 104 & 1040 & 40 \\ 
\hline
\end{tabular}
\caption{The number of layers and heads, the total number of heads, the number of evaluations for each language direction and phenomenon (assuming 10 relevant token-to-token relations), and the number of evaluations when applying our method for each tested model.}
\label{tab:number-evaluations}
\end{table*}

\begin{table*}[!ht]
\centering
\scriptsize
\begin{tabular}{llrrrrrrrrrr}
\hline 
\textbf{Model} & \textbf{Language}
 & \multicolumn{2}{c}{ \textbf{Gender} } & \multicolumn{2}{c}{ \textbf{Auxiliary} } & \multicolumn{2}{c}{ \textbf{Formality} } & \multicolumn{2}{c}{ \textbf{Inflection} } \\ 
 ~ & \textbf{Direction} & Time  & Mem.  & Time & Mem. & Time  & Mem.  & Time  & Mem. \\ 
\hline
 EuroLLM 1.7B Instruct\textsuperscript{\textdagger} & en-de & 0:12 & 9.0 & 0.08 & 13.8 & 0:12 & 11.3 & - & - \\
 ~ & en-fr & 0:12 & 7.8 & 0:22 & 13.0 & 0:11 & 10.1 & - & - \\ 
 ~ & en-pl & 0:24 & 13.2 & 0:21 & 14.1 & 0:17 & 12.6 & 0:13 & 9.9 \\ 
 ~ & en-ru & 0:16 & 14.0 & 0:23 & 13.2 & 0:11 & 12.1 & 0:13 & 10.7 \\ 
\hline
 EuroLLM 9B Instruct\textsuperscript{\textdagger} & en-de & 0:26 & 25.7 & 0:17 & 31.8 & 0:27 & 28.5 & - & - \\ 
 ~ & en-fr & 0:25 & 24.2 & 0:48 & 30.6 & 0:26 & 27.0 & - & - \\ 
 ~ & en-pl & 0:52 & 32.1 & 0:46 & 32.3 & 0:40 & 30.7 & 0:26 & 26.3 \\ 
 ~ & en-ru & 0:36 & 33.7 & 0:49 & 30.9 & 0:26 & 29.9 & 0:27 & 27.7 \\ 
\hline
 Qwen 2.5 1.5B Instruct\textsuperscript{*} & en-de & 0:35 & 9.1 & 0:20 & 14.4 & 0:35 & 12.1 & - & - \\ 
 ~ & en-fr & 0:35 & 7.1 & 0:54 & 13.1 & 0:34 & 10.2 & - & - \\ 
 ~ & en-pl & 1:06 & 15.4 & 0:58 & 16.0 & 0:53 & 15.3 & 0:39 & 11.1 \\ 
 ~ & en-ru & 0:45 & 11.9 & 1:00 & 14.5 & 0:35 & 12.9 & 0.39 & 11.5 \\ 
\hline
 Gemma 3 1B Instruct\textsuperscript{*} & en-de & 0:30 & 11.3 & 0:17 & 19.9 & 0:30 & 15.4 & - & - \\ 
 ~ & en-fr & 0:33 & 8.8 & 0:45 & 18.3 & 0:31 & 13.4 & - & - \\ 
 ~ & en-pl & 1:06 & 19.6 & 0:40 & 23.2 & 0:43 & 18.6 & 0:31 & 13.8 \\ 
 ~ & en-ru & 0:34 & 16.6 & 0:42 & 19.5 & 0:27 & 16.3 & 0:31 & 14.3 \\ 
\hline
\end{tabular}
\caption{The execution time (in hours:minutes) and peak memory (in gigabytes) of a single evaluation for each model and phenomenon. \textsuperscript{\textdagger}NVIDIA H100 80GB. \textsuperscript{*}NVIDIA GeForce RTX~3090 24GB.}
\label{tab:inference-time-memory-requirements}
\end{table*}

\begin{table*}[!ht]
\centering
\scriptsize
\begin{tabular}{llrrrrrrrrrr}
\hline 
\textbf{Model} & \textbf{Language}
 & \multicolumn{2}{c}{ \textbf{Gender} } & \multicolumn{2}{c}{ \textbf{Auxiliary} } & \multicolumn{2}{c}{ \textbf{Formality} } & \multicolumn{2}{c}{ \textbf{Inflection} } \\ 
 ~ & \textbf{Direction} & Time  & Mem.  & Time & Mem. & Time  & Mem.  & Time  & Mem. \\ 
\hline
 EuroLLM 1.7B Instruct\textsuperscript{\textdagger} & en-de & 0:19 & 15.9 & 0:06 & 21.5 & 0:17 & 17.9 & - & - \\ 
 ~ & en-fr & 0:23 & 12.9 & 0:15 & 20.5 & 0:19 & 16.3 & - & - \\ 
 ~ & en-pl & 0:25 & 20.7 & 0:15 & 22.0 & 0:25 & 19.9 & 0:19 & 16.0 \\ 
 ~ & en-ru & 0:18 & 21.9 & 0:16 & 20.7 & 0:16 & 19.1 & 0:18 & 17.1 \\ 
\hline
 EuroLLM 9B Instruct\textsuperscript{\textdagger} & en-de & 0:39 & 58.3 & 0:16 & 72.0 & 0:40 & 62.2 & - & - \\ 
 ~ & en-fr & 0:43 & 51.1 & 0:42 & 69.4 & 0:41 & 59.2 & - & - \\ 
 ~ & en-pl & 0:58 & 70.4 & 0:39 & 73.2 & 0:58 & 68.1 & 0:40 & 58.1 \\ 
 ~ & en-ru & 0:40 & 73.7 & 0:42 & 69.9 & 0:37 & 66.2 & 0:41 & 61.3 \\ 
\hline
 Qwen 2.5 1.5B Instruct\textsuperscript{*} & en-de & 0:55 & 13.4 & 0:19 & 18.8 & 0:55 & 15.8 & - & - \\ 
 ~ & en-fr & 1:00 & 9.3 & 0:50 & 17.0 & 0:56 & 13.2 & - & - \\ 
 ~ & en-pl & 1:19 & 20.2 & 0:51 & 21.0 & 1:20 & 20.1 & 0:58 & 14.5 \\ 
 ~ & en-ru & 0:54 & 15.6 & 0:53 & 18.9 & 0:51 & 16.9 & 0:58 & 15.1 \\ 
\hline
 Gemma 3 1B Instruct\textsuperscript{*} & en-de & 0:59 & 13.3 & 0:17 & 20.3 & 0:30 & 15.4 & - & - \\ 
 ~ & en-fr & 1:08 & 9.4 & 0:45 & 18.8 & 0:58 & 13.9 & - & - \\ 
 ~ & en-pl & 1:15 & 20.3 & 0:44 & 23.2 & 1:16 & 19.1 & 0:58 & 13.4 \\ 
 ~ & en-ru & 0:51 & 16.1 & 0:49 & 20.0 & 0:49 & 16.9 & 0:57 & 14.8 \\ 
\hline
\end{tabular}
\caption{The execution time (in hours:minutes) and peak memory (in gigabytes) of a single gradient calculation run for each model and phenomenon. \textsuperscript{\textdagger}NVIDIA H100 80GB. \textsuperscript{*}NVIDIA GeForce RTX~3090 24GB.}
\label{tab:gradients-time-memory-requirements}
\end{table*}

\section{Details of Head Tuning}
\label{sec:head-tuning-details}

We perform Head Tuning using the ctxPro English-to-German Gender dataset containing 13640 annotated examples. We freeze all parameters of the model apart from the Q and K projections of the selected layer, and calculate the MSE loss between the actual attention scores of the selected head and the attention scores after applying the intervention. Following~\citet{maka-etal-2025-analyzing}, we use the baseline model to calculate the modified attention scores (we load the base model twice: the first one for fine-tuning, the second one for calculating the target attention scores). We used a learning rate of 2e-5 with linear scheduling and no warmup steps, 10 epochs, and a batch size of 4 with 8 gradient accumulation steps. The training took less than one hour on NVIDIA GeForce RTX 3090 24GB.

We evaluate the models on ContraPro contrastive dataset and on 10,000 English-to-German examples sampled from OpenSubtitles 2018 dataset~\citep{lison-etal-2018-opensubtitles2018}. The translation used the greedy decoding (the beam size of 1 and no sampling) with the batch size of 16. The context size was set to 1 and we used the gold (provided) target-side context. We measured BLEU~\citep{papineni-etal-2002-bleu} using the sacreBLEU implementation\footnote{We used \url{https://github.com/mjpost/sacrebleu} with default parameters.}~\citep{post-2018-call} and COMET\footnote{We used the \texttt{Unbabel/wmt22-comet-da} model.}~\citep{rei-etal-2020-comet}.


\section{Details of Indirect Object Identification Experiment}
\label{sec:ioi-details}

In the task of Indirect Object Identification (IOI) \citep{wang2023interpretability} the model has to complete expression in the form "Then, [B] and [A] went to the house. [B] gave a drink to ", where "[A]" and "[B]" are names, and the model should prefer "[A]" over "[B]". In the actual examples, both name placeholders are replaced by popular English names. In the original paper, the names are chosen to be encoded as a single token by the tested model. In case of our model (EuroLLM 1.7B Instruct), the list contained multi-token names, therefore, we removed them when generating the examples. The above example is in the form BABA (corresponding to the order of names). In our experiments we include both BABA and ABBA templates. We generated 10000 unique examples using the code from the paper's repository\footnote{\url{https://github.com/redwoodresearch/Easy-Transformer}}. Due to the simplicity of the task and the near-perfect accuracy exhibited by the tested model, we measure the difference in the logits of the two names from the examples.

We evaluate the model on NVIDIA GeForce RTX 3090 24GB. A single evaluation takes 8 minutes and the registered peak memory on the GPU is 3.1 gigabyte. Calculating gradients lasts 19 minutes with the peak memory of 6.8 gigabytes.
To define the relations of interest, we mark the names in the first part of the IOI sequence as $S_A$ and $S_B$ for names "[A]" and "[B]", respectively, and as $T_A$ and $T_B$ in the second part. We consider the following 14 relations:
\begin{itemize}[topsep=0pt,itemsep=0pt,partopsep=0pt, parsep=0pt, itemindent=15pt, leftmargin=0pt]
        \item $T_{A} \rightarrow S_{A}$,
        \item $T_{A} \rightarrow S_{A+1}$,
        \item $T_{A} \rightarrow S_{B}$,
        \item $T_{A} \rightarrow S_{B+1}$,
        \item $T_{A} \rightarrow T_{B}$,
        \item $T_{A} \rightarrow T_{B+1}$,
        \item $T_{B} \rightarrow S_{A}$,
        \item $T_{B} \rightarrow S_{A+1}$,
        \item $T_{B} \rightarrow S_{B}$,
        \item $T_{B} \rightarrow S_{B+1}$,
        \item $T_{B+1} \rightarrow S_{A}$,
        \item $T_{B+1} \rightarrow S_{A+1}$,
        \item $T_{B+1} \rightarrow S_{B}$,
        \item $T_{B+1} \rightarrow S_{B+1}$.
\end{itemize}
Consequently, we perform 53,760 evaluations modifying heads to $C=0.99$ targeting each head-relation pair. This amounts to 7168 GPU hours, compared to 216 hours for our method when evaluating top 3\% of head-relation pairs, reducing the total time to 3.01\%.




\newpage

\section{Extended Results}
\label{sec:extended-results}

\begin{table*}[!ht]
\centering
\tiny
\begin{tabular}{llrrrrr}
\hline 
\textbf{Model} & \textbf{Language}
 & \multicolumn{1}{l}{ \textbf{Gender} } & \multicolumn{1}{l}{ \textbf{Auxiliary} } & \multicolumn{1}{l}{ \textbf{Formality} } & \multicolumn{1}{l}{ \textbf{Inflection} } \\ 
 ~ & \textbf{Direction} & ~  & ~  & ~  & ~  \\ 
\hline
 EuroLLM 1.7B Instruct & en-de & $84.8^{+0.8}_{-0.8}$& $88.0^{+1.4}_{-1.5}$& $88.9^{+0.8}_{-0.8}$& -  \\ 
 ~ & en-fr & $94.6^{+0.5}_{-0.5}$& $91.3^{+0.8}_{-0.8}$& $91.0^{+0.6}_{-0.7}$& -  \\ 
 ~ & en-pl & $95.4^{+0.4}_{-0.5}$& $87.3^{+0.9}_{-0.9}$& $70.2^{+0.9}_{-1.0}$& $94.0^{+0.5}_{-0.6}$ \\ 
 ~ & en-ru & $90.9^{+0.7}_{-0.7}$& $89.2^{+0.8}_{-0.9}$& $84.0^{+0.9}_{-1.0}$& $96.2^{+0.4}_{-0.5}$ \\ 
\hline
 EuroLLM 9B Instruct & en-de & $87.8^{+0.8}_{-0.8}$& $58.1^{+2.2}_{-2.2}$& $90.9^{+0.7}_{-0.7}$& -  \\ 
 ~ & en-fr & $95.8^{+0.4}_{-0.4}$& $69.0^{+1.3}_{-1.3}$& $65.9^{+1.1}_{-1.1}$& -  \\ 
 ~ & en-pl & $94.8^{+0.5}_{-0.5}$& $63.9^{+1.3}_{-1.4}$& $56.1^{+1.0}_{-1.0}$& $81.5^{+0.9}_{-0.9}$ \\ 
 ~ & en-ru & $78.1^{+1.0}_{-1.1}$& $63.7^{+1.3}_{-1.4}$& $61.4^{+1.2}_{-1.3}$& $85.8^{+0.8}_{-0.9}$ \\ 
\hline
 Qwen 2.5 1.5B Instruct & en-de & $69.9^{+1.1}_{-1.1}$& $86.4^{+1.5}_{-1.6}$& $85.9^{+0.9}_{-0.8}$& -  \\ 
 ~ & en-fr & $91.5^{+0.6}_{-0.6}$& $89.1^{+0.9}_{-0.9}$& $91.3^{+0.6}_{-0.7}$& -  \\ 
 ~ & en-pl & $81.8^{+0.8}_{-0.8}$& $74.5^{+1.2}_{-1.2}$& $53.4^{+1.0}_{-1.0}$& $87.9^{+0.8}_{-0.8}$ \\ 
 ~ & en-ru & $85.5^{+0.9}_{-1.0}$& $82.5^{+1.0}_{-1.1}$& $85.8^{+0.9}_{-0.9}$& $95.3^{+0.5}_{-0.5}$ \\ 
\hline
 Gemma 3 1B Instruct & en-de & $73.4^{+1.0}_{-1.0}$& $57.7^{+2.2}_{-2.2}$& $79.9^{+1.0}_{-1.0}$& -  \\ 
 ~ & en-fr & $89.8^{+0.7}_{-0.7}$& $71.3^{+1.3}_{-1.3}$& $79.2^{+0.9}_{-1.0}$& -  \\ 
 ~ & en-pl & $82.3^{+0.8}_{-0.8}$& $50.9^{+1.5}_{-1.4}$& $45.4^{+1.0}_{-1.1}$& $80.2^{+0.9}_{-1.0}$ \\ 
 ~ & en-ru & $76.1^{+1.1}_{-1.1}$& $56.4^{+1.4}_{-1.3}$& $78.2^{+1.1}_{-1.0}$& $88.1^{+0.8}_{-0.7}$ \\ 
\hline
\end{tabular}
\caption{Results in terms of accuracy on the contrastive datasets for different language directions and phenomena of the unmodified models. Superscript and subscript show the 99\% confidence intervals.}
\label{tab:all-base-results}
\end{table*}

\begin{table*}[!ht]
\centering
\scriptsize
\begin{tabular}{lllrrrrrr}
\Xhline{0.7pt}
            \multirow{2}{*}{\textbf{Model}} & \multirow{2}{*}{\textbf{Setting}} & \multirow{2}{*}{\textbf{Loss}} & \multicolumn{2}{c}{\textbf{All}} & \multicolumn{2}{c}{$\mathbf{\Delta A > 0}$} & \multicolumn{2}{c}{$\mathbf{\Delta A \le 0}$}   \\ 
~ & ~ & ~ & \multicolumn{1}{c}{$\mathbf{r}$} & \multicolumn{1}{c}{$\boldsymbol{\rho}$} & \multicolumn{1}{c}{$\mathbf{r}$} & \multicolumn{1}{c}{$\boldsymbol{\rho}$} & \multicolumn{1}{c}{$\mathbf{r}$} & \multicolumn{1}{c}{$\boldsymbol{\rho}$} \\
\Xhline{0.7pt}
\multirowcell{8}[0pt][l]{EuroLLM 1.7B} & \multirowcell{8}[0pt][l]{ContraPro\\Context 5} & Attention & 0.0459 & 0.1756 & 0.1565 & -0.1416 & 0.0187 & \textbf{0.2322} \\
~ & ~ & $\mathcal{L}_{NLL}$ & -0.0364 & -0.2177 & 0.1677 & 0.1659 & -0.0357 & -0.2502 \\
\Xcline{3-9}{0.1pt}
~ & ~ & $\mathcal{L}_{logC}$ & 0.0816 & 0.1217 & 0.5740 & 0.2300 & 0.0219 & 0.0855 \\
~ & ~ & $\mathcal{L}_{logTC}$ & 0.0782 & 0.0363 & 0.6267 & 0.3796 & -0.0140 & -0.0764 \\
~ & ~ & $\mathcal{L}_{TC}$ & 0.2181 & 0.1178 & 0.6492 & 0.4480 & 0.1760 & -0.0049 \\
\Xcline{3-9}{0.1pt}
~ & ~ & $\mathcal{L}_{logMM}$ & 0.2378 & \textbf{0.2008} & 0.6544 & 0.4060 & 0.2156 & 0.1232 \\
~ & ~ & $\mathcal{L}_{logTMM}$ & 0.2324 & 0.1469 & 0.6747 & 0.4972 & 0.2118 & 0.0183 \\
~ & ~ & $\mathcal{L}_{TMM}$ & \textbf{0.2768} & 0.1711 & \textbf{0.7028} & \textbf{0.5268} & \textbf{0.2737} & 0.0423 \\
\Xhline{0.7pt}
\multirowcell{8}[0pt][l]{EuroLLM 1.7B} & \multirowcell{8}[0pt][l]{ContraPro\\Context 1} & Attention & 0.0329 & 0.0650 & 0.1368 & -0.1433 & 0.0244 & 0.1241 \\
~ & ~ & $\mathcal{L}_{NLL}$ & -0.0086 & -0.1579 & 0.0722 & 0.1342 & -0.0171 & -0.2014 \\
\Xcline{3-9}{0.1pt}
~ & ~ & $\mathcal{L}_{logC}$ & 0.2221 & 0.2363 & 0.6180 & 0.3536 & 0.1972 & 0.1640 \\
~ & ~ & $\mathcal{L}_{logTC}$ & 0.3192 & 0.2869 & \textbf{0.6621} & 0.5059 & 0.3379 & 0.1494 \\
~ & ~ & $\mathcal{L}_{TC}$ & 0.3017 & 0.2902 & 0.6509 & 0.4798 & 0.3041 & 0.1547 \\
\Xcline{3-9}{0.1pt}
~ & ~ & $\mathcal{L}_{logMM}$ & 0.2730 & 0.2284 & 0.6194 & 0.4043 & 0.2738 & 0.1382 \\
~ & ~ & $\mathcal{L}_{logTMM}$ & 0.2974 & 0.2790 & 0.6451 & 0.5151 & 0.3272 & 0.1366 \\
~ & ~ & $\mathcal{L}_{TMM}$ & \textbf{0.3492} & \textbf{0.3330} & 0.6217 & \textbf{0.5463} & \textbf{0.4033} & \textbf{0.1924} \\
\Xhline{0.7pt}
\multirowcell{8}[0pt][l]{EuroLLM 9B*} & \multirowcell{8}[0pt][l]{ContraPro} & Attention & 0.0248 & -0.0954 & 0.3834 & 0.0298 & -0.0405 & -0.1075 \\
~ & ~ & $\mathcal{L}_{NLL}$ & -0.0432 & -0.2059 & 0.0254 & -0.0005 & -0.0183 & -0.1865 \\
\Xcline{3-9}{0.1pt}
~ & ~ & $\mathcal{L}_{logC}$ & 0.1904 & 0.0380 & 0.4259 & 0.1751 & 0.1827 & -0.0020 \\
~ & ~ & $\mathcal{L}_{logTC}$ & 0.2308 & 0.1680 & 0.4639 & 0.2495 & 0.2383 & 0.0847 \\
~ & ~ & $\mathcal{L}_{TC}$ & 0.2817 & 0.1927 & 0.4689 & 0.2550 & 0.3124 & 0.1142 \\
\Xcline{3-9}{0.1pt}
~ & ~ & $\mathcal{L}_{logMM}$ & 0.3597 & 0.1671 & 0.6416 & 0.2758 & 0.4185 & 0.0807 \\
~ & ~ & $\mathcal{L}_{logTMM}$ & 0.3403 & 0.2316 & 0.6278 & 0.3064 & 0.3975 & 0.1248 \\
~ & ~ & $\mathcal{L}_{TMM}$ & \textbf{0.4234} & \textbf{0.3142} & \textbf{0.7054} & \textbf{0.3618} & \textbf{0.4897} & \textbf{0.1921} \\
\Xhline{0.7pt}
\multirowcell{8}[0pt][l]{Qwen 1.5B} & \multirowcell{8}[0pt][l]{ContraPro} & Attention & 0.0528 & 0.0783 & 0.0996 & -0.1318 & 0.0480 & 0.0737 \\
~ & ~ & $\mathcal{L}_{NLL}$ & -0.0602 & -0.0897 & 0.1302 & 0.1592 & -0.0785 & -0.1658 \\
\Xcline{3-9}{0.1pt}
~ & ~ & $\mathcal{L}_{logC}$ & 0.0970 & 0.2573 & 0.6234 & 0.3410 & 0.0419 & 0.1098 \\
~ & ~ & $\mathcal{L}_{logTC}$ & 0.1148 & 0.2395 & 0.7061 & 0.4574 & 0.0686 & 0.0505 \\
~ & ~ & $\mathcal{L}_{TC}$ & 0.1225 & 0.2311 & 0.6950 & 0.4366 & 0.0834 & 0.0504 \\
\cline{3-9}
~ & ~ & $\mathcal{L}_{logMM}$ & 0.1111 & 0.2888 & 0.7113 & 0.4125 & 0.0549 & \textbf{0.1165} \\
~ & ~ & $\mathcal{L}_{logTMM}$ & 0.1214 & 0.2664 & 0.7333 & 0.4855 & 0.0754 & 0.0654 \\
~ & ~ & $\mathcal{L}_{TMM}$ & \textbf{0.1332} & \textbf{0.3108} & \textbf{0.7470} & \textbf{0.4939} & \textbf{0.0913} & 0.1139 \\
\Xhline{0.7pt}
\multirowcell{8}[0pt][l]{Qwen 1.5B} & \multirowcell{8}[0pt][l]{LCPT} & Attention & 0.0496 & -0.0205 & 0.1790 & -0.0810 & \textbf{0.0685} & 0.0906 \\
~ & ~ & $\mathcal{L}_{NLL}$ & -0.0480 & 0.0527 & 0.1578 & 0.2352 & -0.0704 & -0.2398 \\
\Xcline{3-9}{0.1pt}
~ & ~ & $\mathcal{L}_{logC}$ & 0.0617 & 0.2465 & 0.7668 & 0.2906 & 0.0265 & \textbf{0.1081} \\
~ & ~ & $\mathcal{L}_{logTC}$ & \textbf{0.0708} & \textbf{0.2856} & 0.7743 & 0.3356 & 0.0449 & 0.0973 \\
~ & ~ & $\mathcal{L}_{TC}$ & 0.0600 & 0.2806 & 0.8038 & 0.3811 & 0.0118 & 0.0577 \\
\Xcline{3-9}{0.1pt}
~ & ~ & $\mathcal{L}_{logMM}$ & 0.0469 & 0.2144 & 0.8239 & 0.3887 & -0.0067 & -0.0255 \\
~ & ~ & $\mathcal{L}_{logTMM}$ & 0.0575 & 0.2319 & \textbf{0.8326} & \textbf{0.4271} & 0.0171 & -0.0441 \\
~ & ~ & $\mathcal{L}_{TMM}$ & 0.0459 & 0.2826 & 0.8307 & 0.3587 & -0.0187 & 0.0737 \\
\Xhline{0.7pt}
\multirowcell{8}[0pt][l]{Qwen 1.5B} & \multirowcell{8}[0pt][l]{Auxiliary} & Attention & -0.0979 & 0.0491 & 0.0072 & 0.0638 & -0.1230 & -0.0053 \\
~ & ~ & $\mathcal{L}_{NLL}$ & -0.0303 & 0.0386 & 0.6507 & 0.4656 & -0.0885 & -0.2675 \\
\Xcline{3-9}{0.1pt}
~ & ~ & $\mathcal{L}_{logC}$ & 0.0687 & \textbf{0.2875} & 0.8459 & 0.5902 & 0.0148 & \textbf{0.0312} \\
~ & ~ & $\mathcal{L}_{logTC}$ & 0.0718 & 0.2449 & 0.8707 & \textbf{0.6149} & 0.0248 & -0.0478 \\
~ & ~ & $\mathcal{L}_{TC}$ & 0.0749 & 0.2366 & 0.8724 & 0.5907 & 0.0286 & -0.0491 \\
\Xcline{3-9}{0.1pt}
~ & ~ & $\mathcal{L}_{logMM}$ & 0.0609 & 0.2445 & 0.8023 & 0.5688 & 0.0076 & -0.0128 \\
~ & ~ & $\mathcal{L}_{logTMM}$ & 0.0697 & 0.2238 & 0.8381 & 0.6068 & 0.0280 & -0.0696 \\
~ & ~ & $\mathcal{L}_{TMM}$ & \textbf{0.0820} & 0.2430 & \textbf{0.8803} & 0.5984 & \textbf{0.0432} & -0.0450 \\
\Xhline{0.7pt}
\multirowcell{8}[0pt][l]{Gemma 1B} & \multirowcell{8}[0pt][l]{ContraPro} & Attention & 0.0502 & 0.0908 & 0.1447 & -0.0356 & 0.0006 & 0.0293 \\
~ & ~ & $\mathcal{L}_{NLL}$ & -0.1424 & -0.2230 & 0.2575 & 0.3173 & -0.1423 & -0.2481 \\
\Xcline{3-9}{0.1pt}
~ & ~ & $\mathcal{L}_{logC}$ & 0.2814 & 0.2022 & 0.5688 & 0.3447 & 0.2779 & 0.1632 \\
~ & ~ & $\mathcal{L}_{logTC}$ & 0.3139 & 0.2058 & 0.6048 & 0.4330 & 0.3464 & 0.1534 \\
~ & ~ & $\mathcal{L}_{TC}$ & 0.3886 & 0.2508 & 0.6477 & 0.4579 & 0.4340 & 0.1927 \\
\Xcline{3-9}{0.1pt}
~ & ~ & $\mathcal{L}_{logMM}$ & 0.3336 & 0.2274 & 0.6206 & 0.3762 & 0.3652 & 0.1765 \\
~ & ~ & $\mathcal{L}_{logTMM}$ & 0.3375 & 0.2366 & 0.6271 & 0.4376 & 0.3778 & 0.1746 \\
~ & ~ & $\mathcal{L}_{TMM}$ & \textbf{0.4155} & \textbf{0.2892} & \textbf{0.6712} & \textbf{0.4698} & \textbf{0.4613} & \textbf{0.2113} \\
\Xhline{0.7pt}
\end{tabular}
\caption{Pearson $r$ and Spearman $\rho$ correlation coefficients between accumulated gradients and difference in accuracy $\Delta A$ for different sets of head-relation pairs (all, $\Delta A > 0$, and $\Delta \le 0$) when attention scores are modified to $C=0.99$ for all relations of interest for all models and experimental settings (we also include the results when using average attention scores instead of accumulated gradients). *EuroLLM 9B was evaluated on the limited set of token-to-token relations.}
\label{tab:full-correlation-results}
\end{table*}

\begin{figure*}[!ht]
\center{}
    \begin{subfigure}{0.39\linewidth}
        \includegraphics[width=1\linewidth]{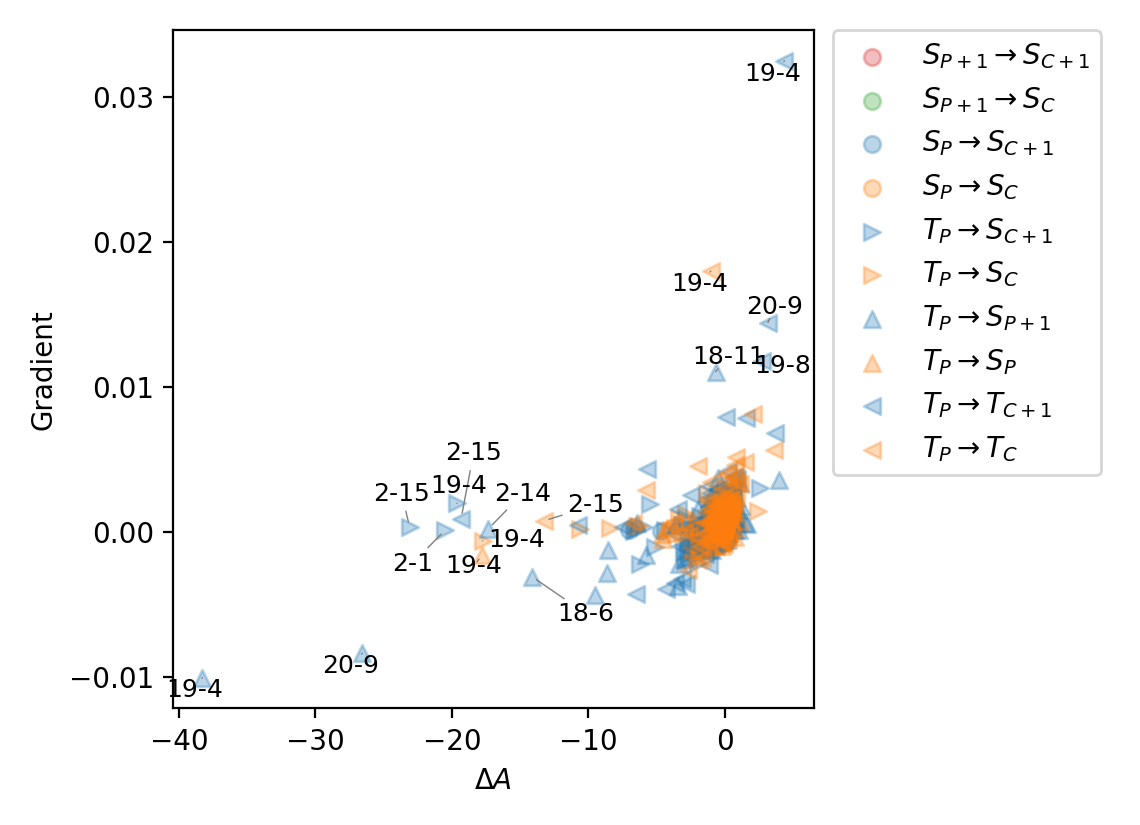}
        \label{fig:visualization-en-de-4-grouping}
    \end{subfigure}
    %
    \begin{subfigure}{0.29\linewidth}
        \includegraphics[width=1\linewidth]{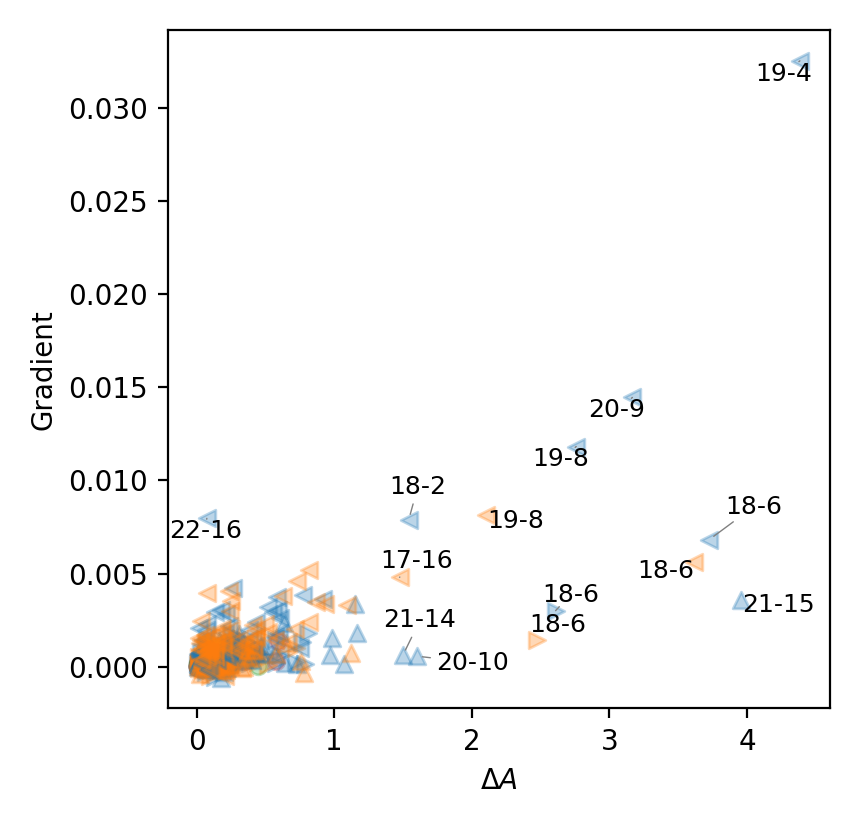}
        \label{fig:visualization-en-de-4-selecting}
    \end{subfigure}\hfill
    \caption{The relation between the accumulated gradients based on the Token-level Max-Margin Loss $\mathcal{L}_{TMM}$ and the change in the accuracy $\Delta A$ when modified to $C=0.99$ for all heads and different relations for EuroLLM 1.7B Instruct on ContraPro with context size of 5. Left figure shows all results, while the right figure shows only the head-relation pairs for the values of $\Delta A > 0$.}
    \label{fig:gradient-accuracy-comparison}
\end{figure*}

\begin{figure*}[!ht]
    \centering
    \includegraphics[width=0.88\linewidth]{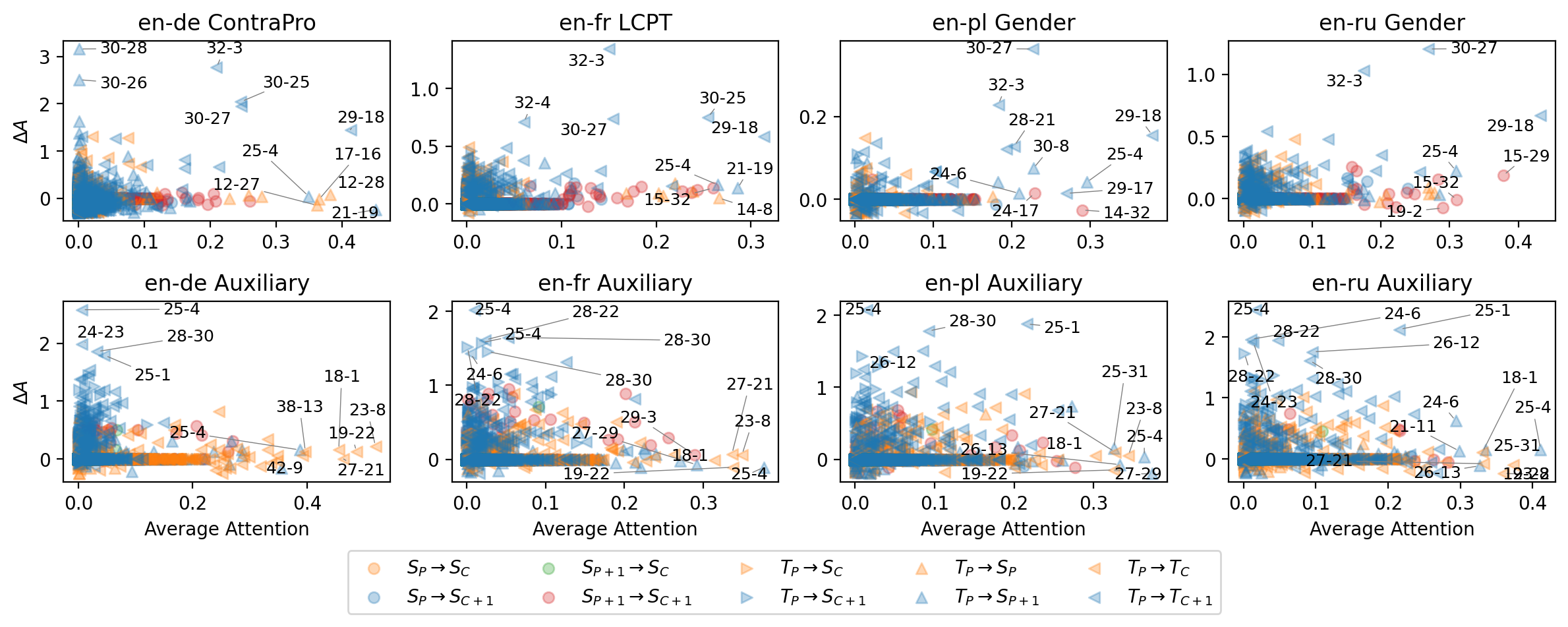}
    \caption{The results in terms of average attention scores (x axes) and accuracy increase $\Delta A$ (y axes) for head-relation pairs for \textbf{EuroLLM 9B Instruct} on all tested phenomena and language directions. To improve clarity, we omitted the head-relation pairs with a large negative $\Delta A$.}
    \label{fig:eurollm-9b-attention-accuracy}
\end{figure*}

\begin{figure*}[!ht]
    \centering
    \includegraphics[width=0.88\linewidth]{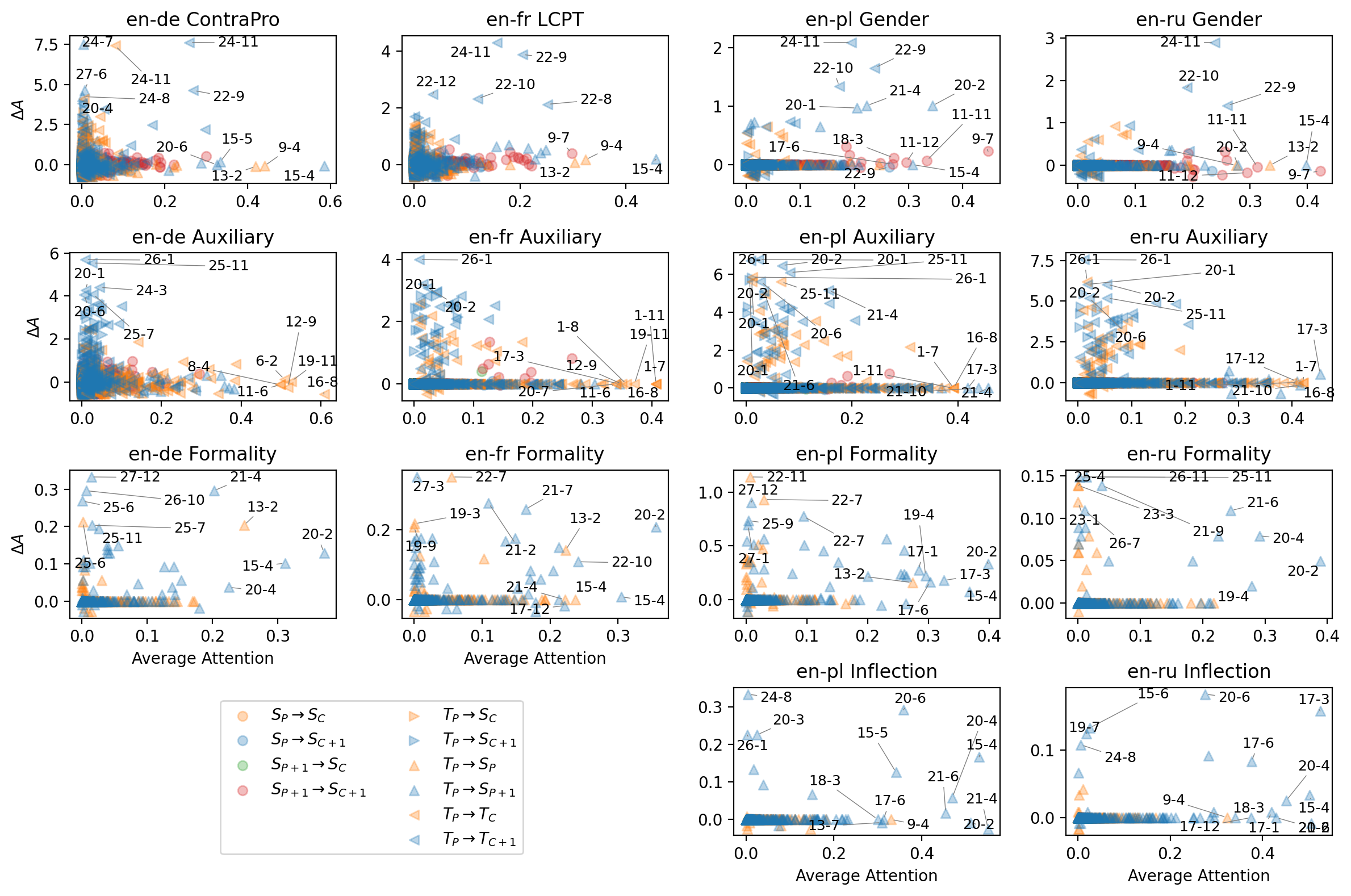}
    \caption{The results in terms of average attention scores (x axes) and accuracy increase $\Delta A$ (y axes) for head-relation pairs for \textbf{Qwen 2.5 1.5B Instruct} on all tested phenomena and language directions. To improve clarity, we omitted the head-relation pairs with a large negative $\Delta A$.}
    \label{fig:qwen-attention-accuracy}
\end{figure*}

\begin{figure*}[!ht]
    \centering
    \includegraphics[width=0.88\linewidth]{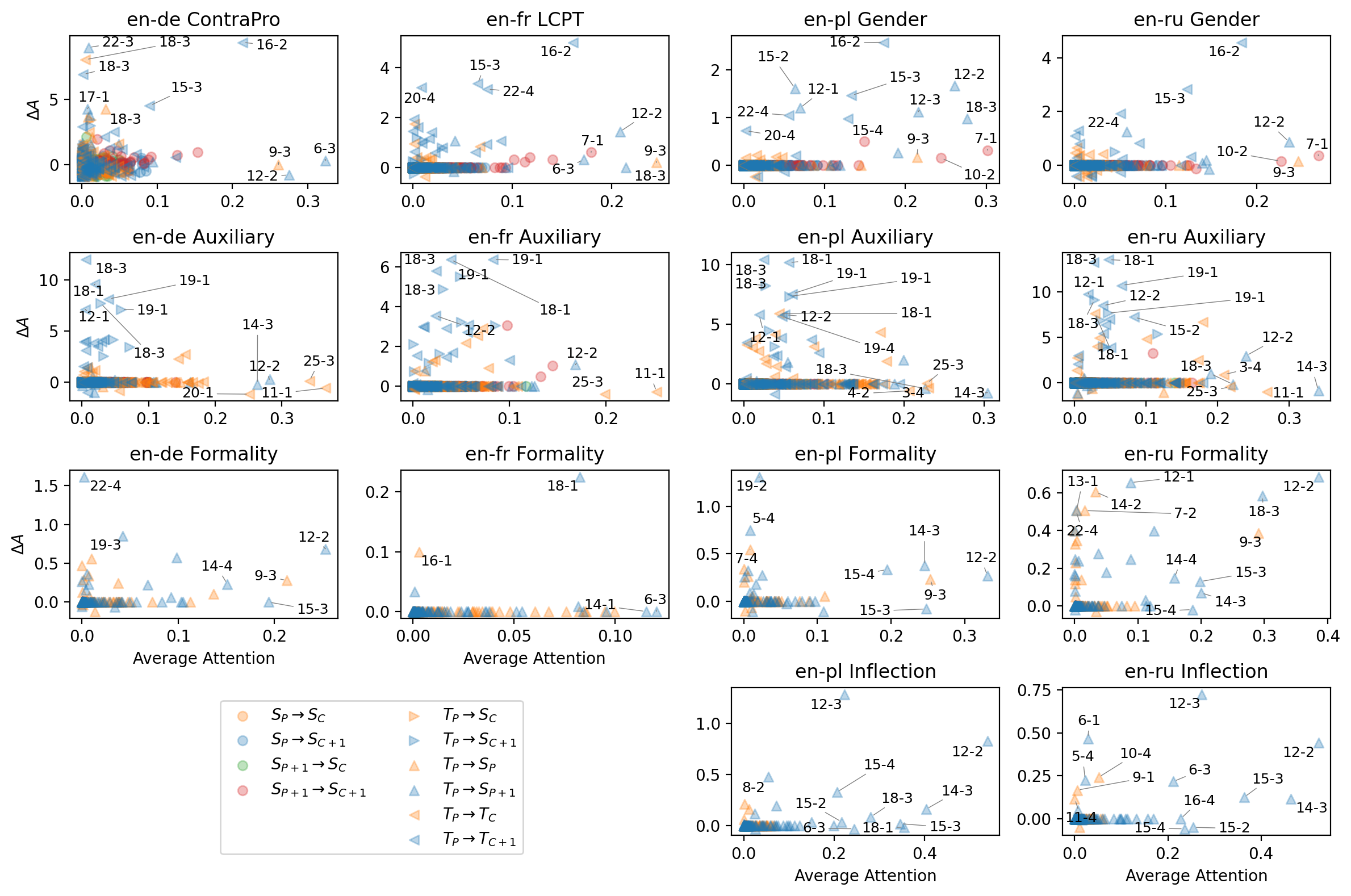}
    \caption{The results in terms of average attention scores (x axes) and accuracy increase $\Delta A$ (y axes) for head-relation pairs for \textbf{Gemma 3 1B Instruct} on all tested phenomena and language directions. To improve clarity, we omitted the head-relation pairs with a large negative $\Delta A$.}
    \label{fig:gemma-attention-accuracy}
\end{figure*}

\end{document}